\documentclass[lettersize,journal]{IEEEtran}
\usepackage{amsmath,amsfonts}
\usepackage[lined]{algorithm2e}

\usepackage{algorithm}
\usepackage{float}
\usepackage{multicol}
\usepackage{array}
\usepackage{subfig}
\usepackage{textcomp}
\usepackage{stfloats}
\usepackage{url}
\usepackage{verbatim}
\usepackage{graphicx}
\usepackage{cite}
\usepackage{amssymb}
\usepackage{bm}
\usepackage{amsmath}
\usepackage{xfrac}
\usepackage{breqn}
\usepackage{amsthm}
\usepackage{bbm}
\usepackage{booktabs}
\usepackage{multirow}
\usepackage{multicol}
\usepackage{tabularx}
\usepackage{makecell}
\usepackage{caption}
\usepackage{xcolor}
\usepackage[colorlinks=true,urlcolor=blue,linkbordercolor=white]{hyperref}

\usepackage{appendix}
\newtheorem{theorem}{Proposition}
\newtheorem{remark}{Remark}
\begin{document}

\title{MRIC: Model-Based Reinforcement-Imitation Learning with Mixture-of-Codebooks for Autonomous Driving Simulation}

\author{Baotian He, Yibing Li
\thanks{Baotian He and Yibing Li are with the State Key Laboratory of Intelligent Green Vehicle and Mobility, School of Vehicle and Mobility, Tsinghua University, Beijing 100084, China. (e-mail: hbt18@mails.tsinghua.edu.cn; liyb@tsinghua.edu.cn)
}
}

\markboth{Journal of \LaTeX\ Class Files,~Vol.~14, No.~8, August~2021}%
{Shell \MakeLowercase{\textit{et al.}}: A Sample Article Using IEEEtran.cls for IEEE Journals}


\maketitle

\begin{abstract}
Accurately simulating diverse behaviors of heterogeneous agents in various scenarios is fundamental to autonomous driving simulation.
This task is challenging due to the multi-modality of behavior distribution, the high-dimensionality of driving scenarios, distribution shift, and incomplete information.
Our first insight is to leverage state-matching through differentiable simulation to provide meaningful learning signals and achieve efficient credit assignment for the policy.
This is demonstrated by revealing the existence of gradient highways and inter-agent gradient pathways.
However, the issues of gradient explosion and weak supervision in low-density regions are discovered.
Our second insight is that these issues can be addressed by applying dual policy regularizations to narrow the function space.
Further considering diversity, our third insight is that the behaviors of heterogeneous agents in the dataset can be effectively compressed as a series of prototype vectors for retrieval.
These lead to our model-based reinforcement-imitation learning framework with temporally abstracted mixture-of-codebooks (MRIC).
MRIC introduces the open-loop model-based imitation learning regularization to stabilize training, and model-based reinforcement learning (RL) regularization to inject domain knowledge. 
The RL regularization involves differentiable Minkowski-difference-based collision avoidance and projection-based on-road and traffic rule compliance rewards.
A dynamic multiplier mechanism is further proposed to eliminate the interference from the regularizations while ensuring their effectiveness.
Experimental results using the large-scale Waymo open motion dataset show that MRIC outperforms state-of-the-art baselines on diversity, behavioral realism, and distributional realism, with large margins on some key metrics (e.g., collision rate, minSADE, and time-to-collision JSD).
\end{abstract}


\section{Introduction}

\begin{figure}[!t]
\centering
\hfil
\subfloat[\href{https://drive.google.com/file/d/12xizenN-GRwB4-3uE7MOD-kBnDINeVA_/view?usp=sharing}{Pure IL variant}]{\includegraphics[]{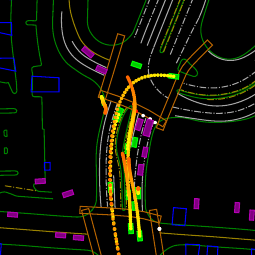}}\label{}\hfil
\subfloat[\href{https://drive.google.com/file/d/1PEpa85gYHd1BPcwn2zAFY4ziArPVXqjO/view?usp=sharing}{\texttt{MRIC}}]{\includegraphics[]{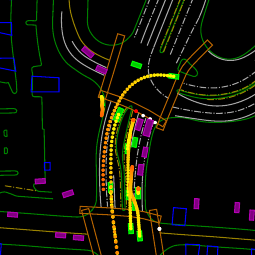}}\label{}\hfil\\
\vspace{-0.2cm}
\hfil
\subfloat[]{\includegraphics[]{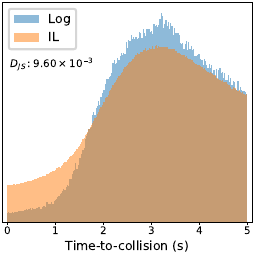}}\hfil
\subfloat[]{\includegraphics[]{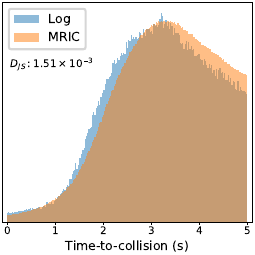}}\hfil
\caption{Comparison of generated behaviors and state-visitation distribution between pure IL variant and the proposed \texttt{MRIC} framework. (a, b): Lime green agents are simulated by the policy. Dynamic visualizations are provided in the captions. (c, d): The y-axis represents frequency. State-matching via closed-loop differentiable simulation provides meaningful learning signals and achieves efficient credit assignment~\cite{minsky1961steps}. However, the policy within this paradigm is scarcely constrained in low-density trajectory regions.
\texttt{MRIC} introduces the model-based RL regularization to provide domain knowledge in subregions not covered by the data distribution, thereby effectively improving behavioral and distributional realism.
}
\label{fig:ablation_framework}
\end{figure}

Autonomous driving (AD) is widely recognized as possessing significant potential to revolutionize mobility and logistics.
With the emergence of deep-learning-based artificial intelligence, AD technologies have achieved encouraging progress in recent years.
However, AD systems capable of handling a wide range of traffic scenarios have yet to become commercially available so far.
One significant reason is the insufficient capability of AD systems to accurately understand and mimic human decisions and behaviors.
AD simulation, which aims to facilitate developing and deploying AD systems by simulating the physical world, is receiving increasing attention.
Nevertheless, human behavior in the real world is characterized by high levels of randomness with complex interactions.
Therefore, how to accurately simulate the diverse behaviors of different types of traffic participants in distinct contexts remains an open problem in the field.

Both academia and industry have made great efforts to construct integrated simulators for training and validating AD systems, such as CARLA~\cite{dosovitskiy2017carla}, Waymo's SimulationCity~\cite{simulation_city}, Wayve's Ghost Gym~\cite{ghost_gym}, NVIDIA DRIVE Sim~\cite{nvidia_drive_sim} and Waabi World~\cite{waabi_2024}.
However, while these simulators mainly focus on generating realistic sensor data, little attention has been paid to the realism of traffic agents' dynamic behaviors.
Traditional traffic simulators like SUMO~\cite{lopez2018microscopic} and VISSIM~\cite{fellendorf2010microscopic} typically employ physics-driven models such as IDM~\cite{treiber2000congested} and MOBIL~\cite{kesting2007general}. However, these models are constrained in their ability to model human driving behaviors due to overly strict assumptions and limited expressiveness.
On the other hand, learning-based methods have recently become dominant in AD simulation, demonstrating superiority over traditional traffic models~\cite{suo2021trafficsim, yan2023learning}.
Among them, behavior cloning (BC) based methods~\cite{bergamini2021simnet, sun2022intersim} suffer from covariate shift~\cite{ross2011reduction} and causal confusion~\cite{de2019causal} issues. However, its open-loop training paradigm offers high training stability, and its performance scales with the size of the dataset and model capability.
In closed-loop methods, generative adversarial imitation learning (GAIL) has been employed in~\cite{kuefler2017imitating, bhattacharyya2022modeling}.
However, GAIL is notorious for unstable training and model collapse issues~\cite{goodfellow2016nips}, as well as potentially meaningless discriminator representation for challenging tasks~\cite{huang2023policy}.
Recently proposed state-matching via differentiable simulation (DS) obtains promising results~\cite{suo2021trafficsim, scibior2021imagining}.
However, this paradigm is empirically difficult to achieve stable training.
Besides, as shown in Fig.~\ref{fig:ablation_framework}, pure imitation learning (IL) methods have a higher probability of producing unrealistic behaviors and lower distributional realism with a lack of safety awareness.
Considering diversity, explicit methods that construct anchors, such as route~\cite{igl2022symphony} and endpoint~\cite{xu2023bits}, may be limited by the introduction of inductive bias that affect their capability to capture the full behavior spectrum.
Implicit methods~\cite{suo2021trafficsim, scibior2021imagining} that encode modal information as latent variables currently face restrictions due to a suboptimal latent space, such as prior holes~\cite{rezende2018taming} and posterior collapse~\cite{lucas2019don} issues.
Furthermore, most previous works focus solely on specific types of agents, such as vehicles.

The goal of this study is to simulate diverse behaviors of heterogeneous agents in various scenarios with both behavioral and distributional realism. 
Towards this objective, there are four main challenges, including \textit{multi-modality}, \textit{curse of dimensionality}, \textit{distribution shift} and \textit{incompleteness of information}, as detailed in Sec.~\ref{sec:challenges}.
To address the above challenges and existing research gap, we propose \texttt{MRIC}, a holistic \textbf{M}odel-based \textbf{R}einforcement-\textbf{I}mitation learning framework with mixture-of-\textbf{C}odebooks.
Specifically, by deriving the analytic gradient expressions, we first reveal the the existence of gradient highways~\cite{he2016deep, he2016identity} and cross-step inter-agent gradient pathways in closed-loop differentiable simulation, which theoretically demonstrates this paradigm's superiority in credit assignment~\cite{minsky1961steps}.
By analyzing  gradient norm's upper bound and state-matching objective's nature, we also highlight the issues of gradient explosion and weak supervision in low-density regions.
Based on these findings, \texttt{MRIC}'s basic idea is to address the above issues by applying two policy regularizations, while leveraging state-matching via closed-loop differentiable simulation to provide meaningful learning signals and achieve efficient credit assignment.
Concretely, to stabilize training, we introduce open-loop model-based IL regularization to narrow the function space to the neighborhood of the optimal policy.
To constrain the policy in low-density regions, we propose model-based reinforcement learning (RL) regularization to provide domain knowledge in trajectory regions that are not covered by the data distribution.
In model-based RL, we construct a series of differentiable rewards, including a Minkowski-difference-based collision avoidance reward as well as projection-based on-road and traffic-rule-compliance rewards.
Leveraging the differentiability of the environment and rewards, the multi-agent RL objective is optimized using low-variance analytic gradient for efficient credit assignment.
We further propose a dynamic Lagrangian multiplier mechanism to eliminate the interference of regularizations with the main objective, while simultaneously ensuring their effectiveness.
For designing the latent space, we propose a temporally abstracted mixture-of-codebooks module, which employs a discrete embedding space and a dynamic prior to address the issue of prior holes. The temporal abstraction mechanism can better model the time scale of variations in human behaviors, thus alleviating posterior collapse.
To validate the effectiveness of the proposed framework, we compare \texttt{MRIC} with state-of-the-art learning-based methods on the large-scale Waymo open motion dataset (WOMD).
Experimental results show that \texttt{MRIC} outperforms all the baselines on most metrics, with large margins in some key metrics (e.g., collision rate and minSADE).

The main contributions of this paper are summarized as follows:
\begin{itemize}
\item The existence of gradient highway and cross-step inter-agent gradient pathways in closed-loop differentiable simulation is revealed, theoretically demonstrating its superiority in credit assignment. The issue of gradient explosion is discovered, which explains the training instability phenomenon. The issue of weak supervision in low-density regions, inherent in the state-matching objective, is uncovered, highlighting its bottleneck.
\item An integrated model-based reinforcement-imitation learning framework is proposed, with dual policy regularizations and a dynamic multiplier mechanism. The open-loop model-based IL regularization is introduced to ensure stable training. The model-based RL regularization is introduced to inject domain knowledge, incorporating a proposed differentiable Minkowski-difference-based collision avoidance reward, along with projection-based rewards for on-road and traffic rule compliance. The dynamic multiplier mechanism is proposed to eliminate the interference of the regularizations with the main objective, while ensuring their effectiveness. The proposed framework is able to simulate the diverse behaviors of heterogeneous agents in distinct scenarios with both behavioral and distributional realism.
\item A mixture-of-codebooks module with temporal abstraction mechanism is proposed to compress the diverse behaviors of heterogeneous agents in the dataset as a series of prototype vectors, which effectively alleviates the prior holes and posterior collapse issues.
\end{itemize}

\section{Related works}

\subsection{Open-Loop Behavior Learning}
Behavior learning from human demonstrations is closely related to the tasks of prediction, planning and simulation in autonomous driving.
As the simplest approach, behavior cloning (BC) assumes that state-action pairs in demonstrations are independent identically distributed (i.i.d.), and is formulated as a supervised learning problem.
For example, ALVINN~\cite{pomerleau1988alvinn} and SimNet~\cite{bergamini2021simnet} applied BC to control the behaviors of self-driving vehicle (SDV) and neighboring agents respectively.
The training of BC is stable, and its performance improves with the use of large-scale datasets~\cite{ettinger2021large} and advanced network architectures~\cite{ngiam2021scene, he2022multi}.
However, due to the open-loop training paradigm, BC cannot recover from the out-of-distribution (OOD) states caused by accumulated errors during closed-loop executions~\cite{ross2010efficient}, and is unaware of the causal structure of interactions in the demonstrations~\cite{de2019causal}.
To address these issues, one approach involves data augmentation.
For instance, ChauffeurNet~\cite{bansal2018chauffeurnet} and DAVE-2~\cite{bojarski2016end} synthesized perturbations to the logged state and manually constructed action labels.
Although shown to be effective, the heuristically designed states may be insufficient to encompass all those encountered during execution.
DAgger~\cite{ross2011reduction} instead queried the expert at states visited by the learned policy, which, nevertheless, is inaccessible for a large-scale dataset.
Another approach is to combine learning with planning.
Symphony~\cite{igl2022symphony} performed beam search on the parallel simulations using a learned discriminator to score each state.
BITS~\cite{xu2023bits} proposed a prediction-and-planning module to select action samples that minimize a rule-based cost.
Both methods effectively improve the quality of generated behaviors. However, the planning module may fail when all the samples generated by the policy are of low quality.
Unlike previous works, we address these issues through environment interactions in closed-loop learning, while employing open-loop IL as a policy regularization method to enhance training stability. Additionally, we leverage a differentiable simulator to overcome the i.i.d. assumption of BC, further enhancing its ability to stabilizing training. 

\subsection{Closed-Loop Behavior Learning}
Closed-loop methods unroll the policy within the environment such that the policy's inputs come from a self-induced state distribution.
As a representative approach, adversarial IL minimizes the divergence between the policy's state-action occupancy measure and that of the expert through a minimax game.
In particular, Kuefler \textit{et al.}~\cite{kuefler2017imitating} and Bhattacharyya \textit{et al.}~\cite{bhattacharyya2022modeling} employed generative adversarial imitation learning (GAIL~\cite{ho2016generative}) to simulate human driving behavior in a highway scenario.
Follow-up works such as RTC~\cite{igl2023hierarchical} and Bronstein \textit{et al.}~\cite{bronstein2022hierarchical} applied model-based GAIL (MGAIL~\cite{baram2017end}) to learning the behaviors of neighboring agents and the SDV in an urban scene respectively.
Although it yields good results in controlling a small number of agents, adversarial learning is notorious for its highly unstable training and susceptibility to mode collapse~\cite{goodfellow2016nips}, as well as potentially meaningless discriminator representations~\cite{huang2023policy}.
When a differentiable environment model is available, the distribution matching objective of IL~\cite{ghasemipour2020divergence} can be achieved by directly maximizing the likelihood of the demonstrations.
For example, TrafficSim~\cite{suo2021trafficsim}, ITRA~\cite{scibior2021imagining} and UrbanDriver~\cite{scheel2022urban} learned the control policy by state-matching through differentiable simulation.
PPUU~\cite{henaff2018model} and MILE~\cite{hu2022model} leveraged a learned forward model to train the policy in closed-loop using only offline demonstrations.
Going one step further, we conduct an in-depth analysis of the gradient properties of closed-loop differentiable simulation and the nature of the state-matching objective, highlighting its potential and issues.
We further propose two policy regularizations to address these issues. 

As another typical closed-loop method, RL enables the policy to learn from explicit reward signals via trial and error.
Some recent works have attempted to combine IL with RL, hoping that RL can address safety concerns.
For example, Lu \textit{et al.}~\cite{lu2023imitation} augmented BC with soft actor-critic (SAC~\cite{haarnoja2018soft}) algorithm by alternately optimizing the IL and RL objectives.
Concurrent with our work, RTR~\cite{zhang2023learning} integrated proximal policy optimization (PPO~\cite{schulman2017proximal}) into closed-loop IL through gradient accumulation.
Different from these works, we construct differentiable simulator and rewards to propose a holistic model-based scheme. In this scheme, both the IL objective and RL regularization are optimized by low-variance analytic gradients, ensuring efficient credit assignment. 
Moreover, we propose a dynamic multiplier mechanism to eliminate the interference between the IL objective and RL regularization.

\subsection{Diverse Behavior Learning}
The diversity of human behavior is reflected in the multi-modality of both action distribution and trajectory distribution.
Some works attempted to capture the multi-modal action distribution via expressive policy classes, such as energy-based models~\cite{florence2022implicit} and diffusion models~\cite{pearce2022imitating, chi2023diffusion}.
Although they have promising performance, these models suffer from much slower training and inference speeds compared to a one-step forward policy.
A more efficient way is to employ a hierarchical policy.
A series of works have proposed to condition the policy on heuristically constructed anchors containing modal information, such as routes~\cite{igl2022symphony}, endpoints~\cite{xu2023bits}, maneuvers~\cite{fei2021triple}, relationships~\cite{sun2022intersim} and action clusters~\cite{shafiullah2022behavior}.
However, the introduction of inductive bias may limit the policy's ability to capture the full behavior spectrum of different agent types.
Another series of works instead represent modal information as a latent variable, which is inferred in an unsupervised manner.
For example, VAE-GAIL~\cite{wang2017robust} utilized variational autoencoder (VAE) to infer latent variables through maximizing the likelihood of demonstrations;
InfoGAIL~\cite{li2017infogail}, on the other hand, maximized the mutual information between the latent and the trajectory.
Unlike these works that employ a fixed prior, we use a dynamic prior distribution as the high-level policy, which adjusts the latent distribution based on current observations and significantly alleviates the issue of prior holes.
Other related works in RL community considered learning a continuous embedding space to encode the skill knowledge contained in the datasets for robots~\cite{hausman2018learning, lynch2020learning} and physically simulated characters~\cite{peng2022ase}.
In contrast, our proposed mixture-of-codebooks module adopts a restricted discrete latent space to avoid unseen latent samples during execution, while the mixture design also enables simultaneous modeling of heterogeneous agents.

\begin{figure*}[!t]
\centering
\includegraphics[width=\textwidth]{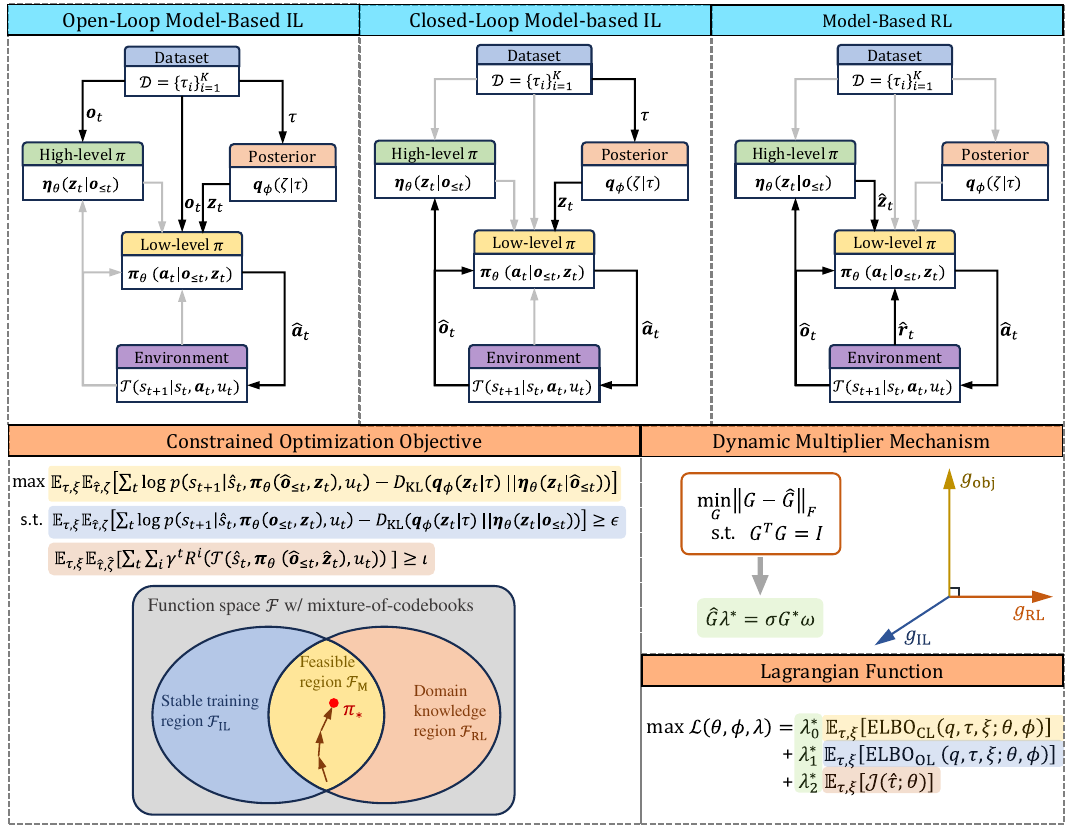}
\caption{Overall framework. The proposed \texttt{MRIC} uses closed-loop model-based imitation learning (IL) as its primary objective. This is supplemented by open-loop model-based IL and model-based reinforcement learning (RL) for regularization. Together, these form a constrained policy optimization problem. The optimal multipliers are obtained through a dynamic multiplier mechanism that solves the multiplier equation. Lastly, the Lagrangian function acts as the dynamic objective for updating the policy at each step.}
\label{fig:framework}
\end{figure*}

\section{Preliminary}

\subsection{Problem Formulation}
In this work, we model the generation process of driving scenarios as a general-sum partially-observable stochastic game (POSG, e.g., \cite{hansen2004dynamic}). A POSG is defined by  $\mathcal{G} = \{\mathcal{S}, \mathcal{A}, \mathcal{O}, P, \{R_i\}_{i=1}^{N}, d_{0}, \gamma\} $. 
$N$ is the number of agents.
$\mathcal{S}$ is the state space. 
$\mathcal{O}$ is the observation space, which is assumed to be shared by all agents.
$\mathcal{A}$ is the union of action spaces for different types of agents.
The joint action and observation spaces are $\mathcal{A}^{N}$ and $\mathcal{O}^{N}$.
The joint actions are sampled from a multi-agent policy $\bm{\pi}(\mathbf{a}|\mathbf{o})$, where $ \mathbf{a} \in \mathcal{A}^{N}$ and $ \mathbf{o} \in \mathcal{O}^{N}$.
$P(s'|s, \mathbf{a})$ is the transition probability function that describes the dynamics of global state, where $ \mathbf{a} \in \mathcal{A}^{N}$ and $s \in \mathcal{S}$.
$\{R_i: \mathcal{S} \times \mathcal{A} \rightarrow \mathbb{R} \}_{i=1}^{N}$ is the set of agents' reward functions, which are not shared with each other. This indicates that agents interact in a mixed cooperative and competitive manner.
$d_{0}=p(s_{0})$ is the distribution of initial state.
$\gamma$ is a discount factor.

A driving scenario can therefore be represented as a trajectory of the POSG, $\tau = \{s_{t}, \mathbf{o}_{t}, \mathbf{a}_{t}, \mathbf{r}_{t} \}_{t=1}^{T} $. At each time step, all agents simultaneously select an action and receive the corresponding observation and reward. The environment accordingly updates the global state given the joint action.
Leveraging the Markov property, the trajectory distribution of naturalistic driving scenario can be factorized as,
\begin{equation}
\label{factorization}
p_{r}(\tau) = p(s_0) \prod_{t=0}^{T-1} p(\mathbf{o}_{t}|s_{\leq t}) \bm{\pi}(\mathbf{a}_{t}|\mathbf{o}_{\leq t}) p(s_{t+1}|s_{t}, \mathbf{a}_{t})
\end{equation}

This work focuses on modelling the multi-agent policy $\bm{\pi}(\mathbf{a}|\mathbf{o})$ for autonomous driving behavior simulation.
Therefore, given a dataset of $K$ sampled trajectories $\mathcal{D} = \{\tau_{i} \}_{i=1}^{K} $, our goal is to learn a parameterized policy $\bm{\pi}_{\theta}$ that minimizes the divergence between the modeled trajectory distribution $p_{\theta}(\tau)$ and data distribution $p_{r}(\tau)$ as,
\begin{equation}
\label{objective}
\min_{\theta} D_{\text{KL}}\left( p_{r}(\tau) \| p_{\theta}(\tau) \right)
\end{equation}
where \small $p_{\theta}(\tau) = p(s_0) \prod_{t=0}^{T-1} p(\mathbf{o}_{t}|s_{\leq t}) \bm{\pi}_{\theta}(\mathbf{a}_{t}|\mathbf{o}_{\leq t}) p(s_{t+1}|s_{t}, \mathbf{a}_{t})$. \normalsize
Here we use the forward Kullback–Leibler (KL) divergence because querying the likelihood from the data distribution in reverse KL divergence is typically intractable.

\subsection{Core Challenges}\label{sec:challenges}
Addressing the aforementioned problem presents significant difficulties. We highlight four key challenges as follows:
\begin{itemize}
\item \textit{Multi-modality}. Due to the different intents and styles of each agent and the different interaction possibilities between agents, both the per-step action distribution and the whole trajectory distribution are highly multi-modal from both marginal and joint perspectives.
\item \textit{Curse of dimensionality}. The complex static and dynamic contexts and numerous neighboring agents make the observation of the policy highly dimensional. This poses great difficulties to credit assignment, and leads to low efficiency in learning implicit driving knowledge, as shown in Fig.~\ref{fig:ablation_framework}.
\item \textit{Distribution shift}. The inevitable prediction errors made by the policy could compound and lead to OOD states for each agent. In interactive scenarios, the error made by one agent could impact and propagate to all other agents as simulation evolves. Besides, the combination explosion of agents' behavioral modes also causes the OOD states.
\item \textit{Incompleteness of information}. Both action and true reward information are inaccessible in naturalistic driving datasets, where only state information can be acquired.
\end{itemize}

\section{Methodology}

\subsection{Overall Architecture}
In this work, we propose to solve the objective~(\ref{objective}) by leveraging differentiable simulation and applying two policy regularizations.
Specifically, it is noted that objective~(\ref{objective}) is equal to $\max_{\theta} \mathbb{E}_{\tau \sim p_{r}(\tau)} [\log p_{\theta}(\tau)] $ ignoring constant terms.
Without a differentiable simulator, this likelihood objective is reduced to behavior cloning of $\max_{\theta} \mathbb{E}_{\tau \sim p_{r}(\tau)}[\sum_{t} \log \bm{\pi}_{\theta}(\bm{a}_{t}|\bm{o}_{\leq t})]$, which suffers from distribution shift~\cite{ross2010efficient} and causal confusion~\cite{de2019causal} issues as well as a lack of action labels.
However, after constructing a differentiable simulator (Sec.~\ref{sec:simulator}) and reparameterized policy (Sec.~\ref{sec:codebook}), objective~(\ref{objective}) is derived as $\max_{\theta} \mathbb{E}_{\tau \sim p_{r}(\tau), \hat{\tau} \sim p_{\theta}(\hat{\tau})}[\sum_{t} \log p(s_{t+1}|\hat{s}_{t}, \bm{\pi}_{\theta}(\bm{\hat{o}}_{\leq t})) ]$ because the policy can be unrolled in the simulator.
In the converted objective, the covariate shift and causal confusion issues are largely alleviated, because the observations come from the self-induced state distribution and temporal credit assignment~\cite{minsky1961steps} is facilitated by differentiable simulation. Besides, the policy can be trained with only logged states. (Sec.~\ref{sec:closed-loop})

Despite these advantages, the closed-loop state-matching objective still has two issues.
First, the training process is unstable. As analyzed in Sec.~\ref{sec:closed-loop}, there is a tendency to experience gradient explosions in closed-loop DS.
Second, the policy is less constrained in low-density regions of the trajectory space. This is due to the zero-avoiding nature of the forward KL divergence~\cite{murphy2012machine}, which causes the learned trajectory distribution $p_{\theta}(\tau)$ to overestimate the support of the data distribution $p_{r}(\tau)$. Consequently, the policy tends to produce diverse but potentially unrealistic behaviors, especially for multi-modal data distribution.
To address the first issue, we propose the open-loop model-based IL regularization $\max_{\theta} \mathbb{E}_{\tau \sim p_{r}(\tau), \hat{\tau} \sim p_{\theta}(\hat{\tau})}[\sum_{t} \log p(s_{t+1}|\hat{s}_{t}, \bm{\pi}_{\theta}(\bm{o}_{\leq t})) ]$ (Sec.~\ref{sec:open-loop}), which constrains the function space $\mathcal{F}$ to a smaller region capable of stable training $\mathcal{F}_{\text{IL}}$.
For the second issue, we introduce the model-based RL regularization $\max_{\theta} \mathbb{E}_{\hat{\tau} \sim p_{\theta}(\hat{\tau})}[\mathcal{J}(\hat{\tau})]$ (Sec.~\ref{sec:RL}), which penalizes the policy for having high probabilities in regions of low real density. Intuitively, this prevents the policy from generating unrealistic behaviors by injecting domain knowledge. 
To achieve efficient credit assignment in high dimensionality, we construct a series of differentiable and accurate rewards (Sec.~\ref{sec:RL}), which are optimized by analytic gradient with low variance.
While the regularizations effectively alleviate the above issues, they can potentially interfere with the optimization of the main objective.
To solve this problem, we further introduce a dynamic multiplier mechanism to eliminate the interference and ensure the effectiveness of the regularizations in Sec.~\ref{MTO}.

Considering the multi-modality challenge, we employ latent variables to represent the behavioral uncertainty, which, however, has potential issues such as prior holes~\cite{rezende2018taming} and posterior collapse~\cite{lucas2019don}. 
Therefore, we propose the mixture-of-codebooks with temporal abstraction mechanism in Sec.~\ref{sec:codebook}, which provides a better-defined function space for policy optimization.
Finally, to deal with the curse of dimensionality, we employ neural networks as the function approximators in Sec.~\ref{sec:net}.

The whole framework is illustrated in Fig.~\ref{fig:framework}. The details of each components are contained in the following sections.

\subsection{Differentiable Simulator}\label{sec:simulator}  
As shown in Eq.~(\ref{factorization}), to generate new scenarios for autonomous driving simulation, we need to model the state transition probability $p(s_{t+1}|s_{t}, \mathbf{a}_{t})$, the observation model $p(\mathbf{o}_{t}|s_{\leq t})$ and the initial state distribution $p(s_0)$ in addition to the multi-agent policy. 
This section describes the construction of these components and derives the general gradient expression of differentiable simulation.

\subsubsection{Components}
For the state transition probability, we abstract the randomness in the distribution as \textit{environmental uncertainty} denoted as $u$, which has semantics like changes of traffic light state. The environmental uncertainty in a driving scenario can therefore be represented as a sequence of latent variables $\xi = \{u_t \}_{t=1}^{T}$.
Since this work focuses on behavior modelling, we model the distribution over environmental latents as an empirical distribution, i.e., $p(\xi) = \frac{1}{K} \sum_{i=1}^{K} \delta (\xi - \xi_{i})$.
Thereby, we can rewrite the state transition probability as a Dirac delta function $p(s_{t+1}|s_{t}, \mathbf{a}_{t}, u_{t}) =  \delta(s_{t+1} - \mathcal{T}(s_{t}, \mathbf{a}_{t}, u_{t}))$, where $\mathcal{T} $ is a deterministic and differentiable state transition function.
Specifically, the information in the global state $s_{t} $ can be decomposed into static map elements $\mathcal{M}$, dynamic map elements $\mathcal{LI}_{t}$ and agent dynamic states $s^{i}_{t}$, i.e., $s_{t} = \{\mathcal{M}, \mathcal{LI}_{t}, s^{1}_{t}, ..., s^{N}_{t} \}$, described in a global coordinate system. 
The static map elements are typically the HD semantic map such as roadgraphs, crosswalks, bikeways. 
The dynamic map elements are status and positions of the traffic lights.
The agent dynamic states consist of position, orientation, and speed, $s^{i}_{t} = (x^{i}_{t}, y^{i}_{t}, \psi^{i}_{t}, v^{i}_{t}) $.
Generally, we have
\begin{align}
s_{t+1}^{i} = f(s_{t}^{i}, a_{t}^{i})
\end{align}
where the action space $\mathcal{A}$ and agent kinematic model $f$ are constructed according to different types of agents. 
For vehicles and cyclists, we adopt the bicycle motion model~\cite{kong2015kinematic} to describe the behavior as,
\begin{align}
x_{t+1}^{i} &= x_{t}^{i} + v_{t}^{i} \text{cos}(\psi_{t}^{i} + \rho_{t}^{i}) \Delta t \\
y_{t+1}^{i} &= y_{t}^{i} + v_{t}^{i} \text{sin}(\psi_{t}^{i} + \rho_{t}^{i}) \Delta t \\
\psi_{t+1}^{i} &= \psi_{t}^{i} + \frac{v_{t}^{i}}{l_{r}^{i}}\text{sin}(\rho_{t}^{i}) \Delta t\\
v_{t+1}^{i} &= v_{t}^{i} + \alpha_{t}^{i} \Delta t \\
\rho_{t}^{i} &= \text{tan}^{-1} \left( \frac{l_{r}^{i}}{l_{f}^{i}+l_{r}^{i}}\text{tan}(\beta_{t}^{i}) \right) 
\end{align}
where $\alpha$ and $\beta$ denote the acceleration and steering angle that constitute the action space $a = (\alpha, \beta) $ for vehicles and cyclists; $l_{r}$ and $l_{f}$ denote the distances from the center to the rear and front axles, which are rewritten as the product of a coefficient and the agent's length, i.e., $l_{r} = c_{r}L$ and $l_{f} = c_{f}L$. The coefficients $c_{f}$ and $c_{r}$ can be solved by grid search or approximated by manually defined values.
The usage of the bicycle motion model allows the modelling in smoother action space instead of state space, which assists in better capturing the behavior characteristics.
For pedestrians, we adopt a delta motion model as follows
\begin{align}
x_{t+1}^{i} &= x_{t}^{i} + \Delta x \\
y_{t+1}^{i} &= y_{t}^{i} + \Delta y \\
\psi_{t+1}^{i} &= \psi_{t}^{i} + \Delta \psi \\
v_{t+1}^{i} &= \sqrt{({\Delta x} / {\Delta t})^{2} + ({\Delta y} / {\Delta t})^{2} }
\end{align}
where $\Delta x$, $\Delta y$ and $\Delta \psi$ are the changes in position and orientation that constitute the action space $a = (\Delta x, \Delta y, \Delta \psi) $ for pedestrians.
Thereby, the Jacobian of the agent dynamic function w.r.t. the state can be derived as
\begin{equation}
\label{jacobian_bicycle}
\frac{\partial f}{\partial s_{t}^{i}} = 
\begin{bmatrix}
1 & 0 & - v_{t}^{i} \text{sin}(\psi_{t}^{i}+\rho_{t}^{i}) \Delta t & \text{cos}(\psi_{t}^{i}+\rho_{t}^{i}) \Delta t \\
0 & 1 & v_{t}^{i} \text{cos}(\psi_{t}^{i}+\rho_{t}^{i}) \Delta t & \text{sin}(\psi_{t}^{i}+\rho_{t}^{i}) \Delta t \\
0 & 0 & 1 & \text{sin}(\rho_{t}^{i}) \Delta t / l_{r}^{i} \\
0 & 0 & 0 & 1
\end{bmatrix}
\end{equation}
for vehicles and cyclists; and 
\begin{equation}
\label{jacobian_delta}
\frac{\partial f}{\partial s_{t}^{i}} = 
\begin{bmatrix}
1 & 0 & 0 & 0 \\
0 & 1 & 0 & 0 \\
0 & 0 & 1 & 0 \\
0 & 0 & 0 & 0
\end{bmatrix}
\end{equation}
for pedestrians.

For the observation probability, we model it as a Dirac delta function, i.e., $p(\mathbf{o}_{t}|s_{\leq t}) = \delta(\mathbf{o}_{t} - \mathcal{OM}(s_{\leq t}))$, where $\mathcal{OM}$ is a deterministic and differentiable observation function.
The partial observability is modeled by extracting a fixed number of scene elements for each agent based on spatial distance.
The extracted elements are then transformed into the local coordinate system of each agent.

For the initial state probability, we model it as the empirical distribution of the sampled initial states, i.e., $p(s_{0}) = \frac{1}{K} \sum_{i=1}^{K} \delta(s_{0} - {s_{0}}_{i}) $.
Considering the large scale of the naturalistic driving dataset, the diversity of the logged initial states can meet the needs of autonomous driving simulation.

\subsubsection{Gradient flow}
To investigate the gradient properties of policy learning through differentiable simulation, we present the following proposition that provides a general gradient expression, decomposed across both time and agent dimensions.
\begin{theorem}
\label{thm:general}
Assume that the observation model and the policy respectively take the current state and observation as inputs, namely $p(\bm{o}_{t}|s_{t})$ and $\bm{\pi}_{\theta}(\bm{a}_{t}|\bm{o}_{t})$, and that the objective function is defined over the state sequence: $J = J(s_{1:T})$.
After unfolding the policy in a differentiable simulator and obtaining the objective value, the general expression of policy gradient is as follows:
\begin{align}
\frac{\partial J}{\partial \theta} &= \sum_{t=1}^{T} \frac{\partial J_{t}}{\partial \theta} \label{objective_decomposition} \\
&= \sum_{t=1}^{T} \sum_{j=1}^{t} \frac{\partial J_{t}}{\partial s_{t}} \frac{\partial s_{t}}{\partial s_{j}} \frac{\partial s_{j}}{\partial \bm{a}_{j-1}} \frac{\partial \bm{a}_{j-1}}{\partial \theta} \label{cross1} \\
&= \sum_{t=1}^{T} \sum_{j=1}^{t} \frac{\partial J_{t}}{\partial s_{t}} \left( \prod_{k=t-1}^{j} \frac{\partial s_{k+1}}{\partial s_{k}} \right) \frac{\partial s_{j}}{\partial \bm{a}_{j-1}} \frac{\partial \bm{a}_{j-1}}{\partial \theta} \label{general_gradient_expression_time} \\
&= \sum_{t=1}^{T} \sum_{j=1}^{t} \sum_{i=1}^{N} \frac{\partial J_{t}}{\partial s_{t}} \left( \prod_{k=t-1}^{j} \frac{\partial s_{k+1}}{\partial s_{k}} \right) \frac{\partial s_{j}}{\partial {a}_{j-1}^{i}} \frac{\partial {a}_{j-1}^{i}}{\partial \theta} \label{general_gradient_expression_agent}
\end{align}
where defining $\prod_{k=t-1}^{t} \frac{\partial s_{k+1}}{\partial s_{k}} = I $.
\end{theorem}
\begin{remark}
The assumption in the above proposition is made to highlight the key properties of the gradient.
The objective function can be state-matching loss (for IL) or cumulative rewards (for RL).
The partial derivatives use numerator-layout notation throughout the paper.
Eq.~(\ref{objective_decomposition}) decomposes the overall objective into the summation of per-step objectives.
Eq.~(\ref{cross1}) and Eq.~(\ref{general_gradient_expression_time}) introduce the cross-step derivative term $\frac{\partial s_{t}}{\partial s_{j}} = \prod_{k} \frac{\partial s_{k+1}}{\partial s_{k}}$, which describes the gradient flow from future time step to preceding ones.
Eq.~(\ref{general_gradient_expression_agent}) further decomposes the gradient across each agent.
\end{remark}
The above proposition serves as a tool for analysing the gradient properties of different policy learning paradigms, which differ in their cross-step derivatives and objective functions.
As shown in Sec.~\ref{sec:closed-loop}, the cross-step derivative plays a critical role in the gradient norm and the agent interaction modelling.

\subsection{Temporally Abstracted Mixture-of-Codebooks}\label{sec:codebook}
Modelling the multi-modal multi-agent policy is the main focus of this paper. In this section, we highlight several key designs that are crucial to the performance of the policy.

\begin{figure}[!t]
\centering
\subfloat[\label{fig:hierarchical_policy}]{\includegraphics[]{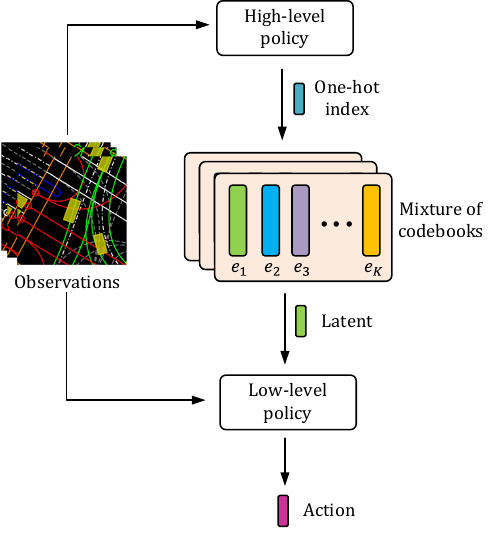}}  \\
\subfloat[\label{fig:temporal_abstraction}]{\includegraphics[]{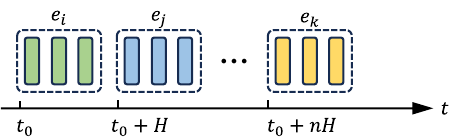}} 
\caption{The proposed temporally abstracted codebook mixture module. (a) Hierarchical policy with mixture-of-codebooks. (b) Temporal abstraction mechanism.}
\label{fig:policy}
\end{figure}

\subsubsection{Hierarchy}
To effectively capture the randomness of the policy, we encode the \textit{behavioral uncertainty} as a latent variable $\bm{z}$, which has semantics like behavioral intent, style, etc. 
Formally, we perform a transformation to the multimodal policy as follows,
\begin{equation}
\bm{p}(\bm{a}_{t}|\bm{o}_{\leq t}) = \bm{\eta} \left(\bm{\pi}^{-1}(\bm{a}_{t}, \bm{o}_{\leq t})|\bm{o}_{\leq t}\right) \left| \frac{\partial \bm{\pi}^{-1}(\bm{a}_{t}, \bm{o}_{\leq t})}{\partial \bm{a}_{t}}  \right|
\end{equation}
where $\bm{z}_{t} \sim \bm{\eta}(\bm{z}|\bm{o}_{\leq t})$ represents the distribution over behavioral latents, and $\bm{a}_{t} = \bm{\pi}(\bm{z}_{t}, \bm{o}_{\leq t})$ is the transformation function that maps latent variables to actions.
The behavioral uncertainty in a driving scenario is thereby represented as a latent sequence of $\zeta = \{ \bm{z}_{t} \}_{t=0}^{T-1} $.
Intuitively, the latent $\bm{z}_{t} $ can be viewed as the specification of the mode component of the original multimodal policy $\bm{p}(\bm{a}_{t}|\bm{o}_{\leq t}) $, and $\bm{\eta}(\bm{z}|\bm{o}_{\leq t}) $ describes the probabilities of different modes. Then the transformation function $\bm{\pi}(\bm{z}_{t}, \bm{o}_{\leq t}) $ generates actions according to the current observation and the selected behavioral mode.
After the transformation, we can adopt Gaussian \cite{kingma2013auto} or categorical \cite{JangGP17} reparameterization for gradient propagation through the policy.

\subsubsection{Temporal abstraction}
Compared to the simulation interval (e.g. $100\mathrm{ms}$), variations in naturalistic human behavior typically occur on a larger time scale (e.g. $5\mathrm{s}$ for lane change \cite{toledo2007modeling}).
Actions within a maneuver are typically close to each other. For instance, the changes in steering angle during a lane change are significantly smaller than those observed when transitioning from lane keeping to lane changing.
Our key insight is that this relative smoothness of actions within a maneuver can be effectively captured by the consistency of the sampled policy's mode component within a time interval.
To impose this prior knowledge on the sampling process, we introduce a temporal abstraction mechanism to the sequence of behavioral latents, i.e.,
\begin{equation}
\begin{cases}
z_{t}^{i} \sim \eta(z|o_{\leq t}^{i}), & \text{if} \ m_{t}^{i} = 1, \\
{z_{t}^{i}} = z_{t-1}^{i}, & \text{otherwise}.
\end{cases}
\end{equation}
where $\bm{m}$ is a binary variable indicating whether to terminate the last latent.
As shown in Fig.~\ref{fig:temporal_abstraction}, we determine $m_{t}^{i}$ at each step using heuristics in this work.
Intuitively, the temporal abstraction mechanism mimics how humans decompose the driving process into a series of maneuvers, enabling short-term planning on these high-level maneuvers instead of each low-level action.

\subsubsection{Mixture-of-codebooks}
When using continuous latent space, the sampled latent variable at inference time may never be seen during training due to the unrestricted latent space. It is shown to degrade the performance of the policy in Sec.~\ref{sec:ablation_latent}.
To alleviate such mismatch, we discretize the latent space into a series of embedding vectors $e_{i} \in \mathbb{R}^{D}, i = 1,\ldots,K$, which form a codebook used in both training and execution.
Considering the distinct behavioral characteristics of different types of agents, we further propose the mixture-of-codebooks module $E = \{E_{i} \in \mathbb{R}^{K \times D} \}$. In this module, each agent type corresponds to a specific codebook, and a hard gating mechanism is employed to selectively activate the appropriate codebook based on the agent's type.
Therefore, the output of the high-level policy becomes the index of the embedding vector in the codebook. The latent variable sent to the low-level policy is selected according to the index as follows,
\begin{equation}
\label{codebook}
z_{t}^{i} = e_{k_{t}^{i}} \quad \text{where}~k_{t}^{i} \sim \eta(k|o_{\leq t}^{i}).
\end{equation}
For brevity, we abbreviate the above indexing process as $\bm{z}_{t} \sim \bm{\eta}(\bm{z}|\bm{o}_{\leq t})$ in the following of the paper.
Intuitively, the learned embedding vectors serve as a series of behavioral prototypes of heterogeneous agents.
So far, we have transformed the problem of modeling a continuous multimodal policy into learning a discrete distribution with a deterministic function, simplifying the task considerably.

\subsubsection{Joint probability decomposition}
\begin{figure}
\centering
\subfloat[]{\includegraphics[]{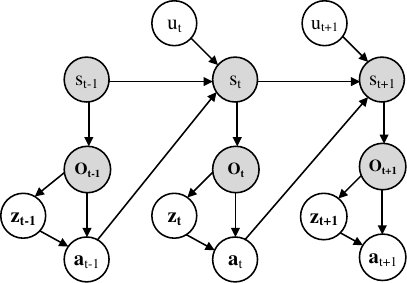}} \\
\subfloat[]{\includegraphics[]{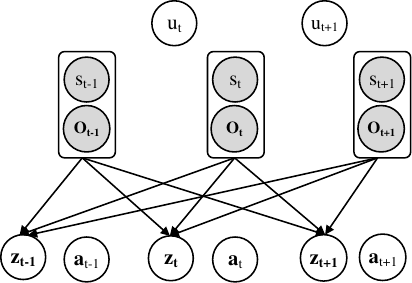}}
\caption{Probabilistic graphical model representation of a driving scenario. The shaded nodes represent observable variables, while the blank nodes represent unobservable variables. Potential links spanning multiple time steps are omitted for clarity. (a) Generation process. (b) Inference process.}
\label{fig:PGM}
\end{figure}

When considering latent variables, the generation process of a driving scenario is illustrated in Fig.~\ref{fig:PGM}. 
The joint probability of a trajectory and the latent sequences is decomposed as,
\begin{equation}
\begin{aligned}
\label{factorization_with_latent}
p(\tau, \zeta, \xi) = p(s_0) \prod_{t=0}^{T-1} & p(\bm{o}_{t}|s_{\leq t}) \bm{\eta}(\bm{z}_{t}|\bm{o}_{\leq t}) \bm{\pi}(\bm{a}_{t}|\bm{o}_{\leq t}, \bm{z}_{t}) \\
&  p(u_{t}|u_{<t}) p(s_{t+1}|s_{t}, \bm{a}_{t}, u_{t})
\end{aligned}
\end{equation}
where $\bm{\pi}$ is a Dirac delta function. $\bm{\eta}$ and $\bm{\pi}$ function as the high- and low-level policies, respectively.

\subsection{Closed-Loop Model-Based Imitation Learning}\label{sec:closed-loop}

\begin{figure*}[!t]
\centering
\subfloat[\label{fig:gradient_bc}]{\includegraphics[]{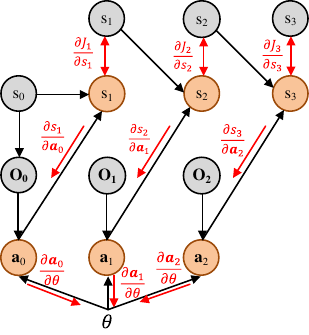}%
\label{gradient_flow_bc}}
\hfil
\subfloat[\label{fig:gradient_olil}]{\includegraphics[]{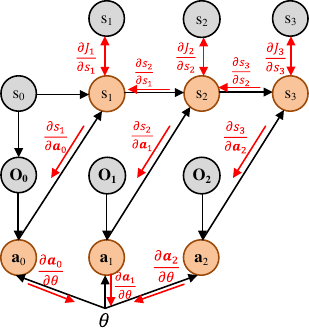}%
\label{gradient_flow_olil}}
\hfil
\subfloat[\label{fig:gradient_clil}]{\includegraphics[]{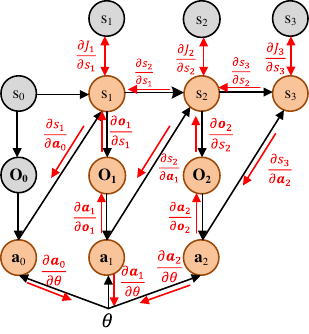}%
\label{gradient_flow_clil}}
\caption{Comparison of gradient flows. (a) Behavior cloning. (b) Open-loop model-based IL via state-matching. (c) Closed-loop model-based IL via state-matching.}
\label{fig_sim}
\end{figure*}

\subsubsection{Objective}
Given a dataset of human trajectories, the goal of closed-loop model-based IL is to maximize the log-likelihood of these trajectories, which however is typically intractable after introducing the latent variables. Therefore, we derive the evidence lower bound (ELBO \cite{kingma2013auto}) of the marginal log-likelihood to serve as the surrogate objective,
\begin{align}
&\mathbb{E}_{\tau \sim \mathcal{D}} \left[\log p_{\theta}(\tau) \right] \nonumber \\
=& \mathbb{E}_{\tau \sim \mathcal{D}} \left[\log \sum_{\zeta} \sum_{\xi} \bm{q}_{\phi}(\zeta|\tau) p(\xi|\tau) \frac{p_{\theta}(\tau, \zeta, \xi)}{\bm{q}_{\phi}(\zeta|\tau) p(\xi|\tau)} \right] \\
\geq& \mathbb{E}_{\tau \sim \mathcal{D}} \left[ \sum_{\zeta} \sum_{\xi} \bm{q}_{\phi}(\zeta|\tau) p(\xi|\tau) \log \frac{p_{\theta}(\tau, \zeta, \xi)}{\bm{q}_{\phi}(\zeta|\tau) p(\xi|\tau)} \right] \label{ELBO} \\
=& \mathbb{E}_{\tau \sim \mathcal{D}, \zeta \sim \bm{q}_{\phi}(\zeta|\tau), \xi \sim p(\xi|\tau)} \left[ \log \frac{p_{\theta}(\tau, \zeta, \xi)}{\bm{q}_{\phi}(\zeta|\tau) p(\xi|\tau)}   \right] 
\end{align}
where $\bm{q}_{\phi}(\zeta|\tau)$ is the variational posterior distribution that approximates the true posterior $\bm{p}(\zeta|\tau) $; the posterior distribution of the environmental latents is the empirical distribution because their values can be directly inferred from the changes between adjacent states in this case, i.e., $p(\xi|\tau) = \frac{1}{K} \sum_{i=1}^{K} \delta (\xi - \xi_{i}) $; and $\mathcal{D} = \frac{1}{K} \sum_{i=1}^{K} \delta (\tau - \tau_{i}) $ denotes the empirical data distribution. 
In principle, the variational distribution  can be decomposed either causally or acausally. For simplicity, we assume conditional independence here, i.e., $\bm{q}_{\phi}(\zeta|\tau) = \prod_{t} \bm{q}_{\phi}(\bm{z}_{t}|\tau) $.

Further substituting the joint probability decomposition Equ.~(\ref{factorization_with_latent}) and ignoring constant terms, we obtain the optimization objective for closed-loop model-based IL,
\begin{align}\label{elbo_cl}
&\mathbb{E}_{\tau \sim \mathcal{D}, \xi \sim p(\xi|\tau)} \left[\text{ELBO-CL}(\bm{q}, \tau, \xi;\theta, \phi)  \right] \nonumber \\
=& \mathbb{E}_{\tau, \xi}  \mathbb{E}_{\zeta \sim \bm{q}_{\phi}(\zeta|\tau), \hat{\tau} \sim p_{\theta}(\hat{\tau}, \zeta, \xi)} [ \sum_{t=0}^{T-1} \log p(s_{t+1}|\hat{s}_{t}, \bm{\pi}_{\theta}( \bm{\hat{o}}_{\leq t}, \bm{z}_{t}), u_{t})   \nonumber \\
-&  D_{\text{KL}} \left(\bm{q}_{\phi}(\bm{z}_{t}|\tau) \| \bm{\eta}_{\theta}(\bm{z}_{t}|\mathbf{\hat{o}}_{\leq t}) \right) ]
\end{align}
where $\hat{\tau} = \{\hat{s}_{t}, \bm{\hat{o}}_{t}\}$ denotes the trajectory generated by the policy.
In this objective, we autoregressively rollout the high- and low-level policies in the differentiable simulator, with behavioral latents inferred by the posterior distribution.
Therefore, during training, the policies' observations stem from a self-induced state distribution, similar to execution, with corresponding action labels implicitly provided by the state-matching objective. This closed-loop training paradigm enables the policies to make corrective actions for deviated observation distribution during execution, which largely alleviates the distribution shift issue~\cite{ross2011reduction}.

\subsubsection{Gradient property}
\begin{figure}[!t]
\centering
\includegraphics{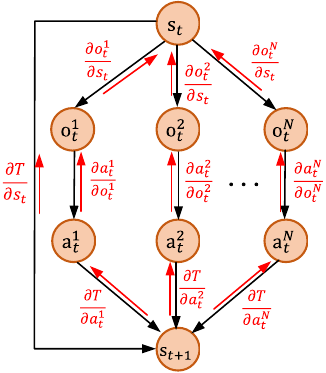}
\caption{Gradient pathways decomposed across agent dimension.}
\label{fig:gradient_interaction}
\end{figure}

We then present the following proposition to reveal the gradient property of policy learning via closed-loop differentiable simulation.
\begin{theorem}
\label{thm:clil}
Given the assumptions stipulated in Proposition 1, the cross-step derivative term in closed-loop imitation learning is derived as follows,
\begin{align}
\frac{\partial s_{t}}{\partial s_{j}} &=  \prod_{k=t-1}^{j} \left( \frac{\partial \mathcal{T}}{\partial s_{k}} + \frac{\partial \mathcal{T}}{\partial \bm{a}_{k}} \frac{\partial \bm{a}_{k}}{\partial \bm{o}_{k}} \frac{\partial \bm{o}_{k}}{\partial s_{k}} \right) \label{csd_branch} \\
&= \prod_{k=t-1}^{j} \frac{\partial \mathcal{T}}{\partial s_{k}} + \sum_{n=1}^{t-j} \left( \frac{\partial \mathcal{T}}{\partial s} \right)^{n} \left( \frac{\partial \mathcal{T}}{\partial \bm{a}} \frac{\partial \bm{a}}{\partial \bm{o}} \frac{\partial \bm{o}}{\partial s}  \right)^{t-j-n} \label{csd_clil}
\end{align}
where $s_{k+1} = \mathcal{T}(s_{k}, \bm{a}_{k})$, $\left( \frac{\partial \mathcal{T}}{\partial s} \right)^{n} \left( \frac{\partial \mathcal{T}}{\partial \bm{a}} \frac{\partial \bm{a}}{\partial \bm{o}} \frac{\partial \bm{o}}{\partial s}  \right)^{t-j-n} $ denotes the partial derivatives that flow through the state branch in the state transition function $n$ times and the action branch $t-j-n$ times. The policy gradient therefore is,
\begin{align}
\label{gradient_cl}
\frac{\partial J}{\partial \theta} =& \sum_{t=1}^{T} \sum_{j=1}^{t} \left[ \frac{\partial J_{t}}{\partial s_{t}} \left(\prod_{k=t-1}^{j} \frac{\partial \mathcal{T}}{\partial s_{k}} \right) \frac{\partial s_{j}}{\partial \bm{a}_{j-1}} \frac{\partial \bm{a}_{j-1}}{\partial \theta} \right.  \\
+& \left. \sum_{n=1}^{t-j} \frac{\partial J_{t}}{\partial s_{t}} \left( \frac{\partial \mathcal{T}}{\partial s} \right)^{n} \left( \frac{\partial \mathcal{T}}{\partial \bm{a}} \frac{\partial \bm{a}}{\partial \bm{o}} \frac{\partial \bm{o}}{\partial s}  \right)^{ t-j-n} \frac{\partial s_{j}}{\partial \bm{a}_{j-1}} \frac{\partial \bm{a}_{j-1}}{\partial \theta} \right]. \nonumber
\end{align}
Further assuming the multi-agent policy is agent-wise factored, i.e., $\bm{\pi}_{\theta}(\bm{a}_{t}|\bm{o}_{t}) = \prod_{i=1}^{N} \pi_{\theta}^{i}(a_{t}^{i}|o_{t}^{i}) $, the cross-step derivative can be expanded as,
\begin{align}
\frac{\partial s_{t}}{\partial s_{j}} &= \prod_{k=t-1}^{j} \left( \frac{\partial \mathcal{T}}{\partial s_{k}} + \sum_{i=1}^{N} \frac{\partial \mathcal{T}}{\partial a_{k}^{i}} \frac{\partial a_{k}^{i}}{\partial o_{k}^{i}} \frac{\partial o_{k}^{i}}{\partial s_{k}} \right) \\
&= \sum_{n=0}^{t-j} \sum_{i_{1}=1}^{N} \cdots \sum_{i_{n}=1}^{N} \left( \frac{\partial \mathcal{T}}{\partial s} \right)^{t-j-n} \left( \frac{\partial \mathcal{T}}{\partial a^{i}} \frac{\partial a^{i}}{\partial o^{i}} \frac{\partial o^{i}}{\partial s}  \right)^{n} \label{gradient_agent_interaction}
\end{align}
The upper bound of cross-step derivative's norm is derived as,
\begin{align}
\left\| \prod_{k=t-1}^{j} \frac{\partial s_{k+1}}{\partial s_{k}} \right\| &\leq \prod_{k=t-1}^{j} \left\| \frac{\partial \mathcal{T}}{\partial s_{k}} + \frac{\partial \mathcal{T}}{\partial \bm{a}_{k}} \frac{\partial \bm{a}_{k}}{\partial \bm{o}_{k}} \frac{\partial \bm{o}_{k}}{\partial s_{k}} \right\| \\
&\leq \prod_{k=t-1}^{j} \left\| \frac{\partial \mathcal{T}}{\partial s_{k}}  \right\| + \left\| \frac{\partial \mathcal{T}}{\partial \bm{a}_{k}} \frac{\partial \bm{a}_{k}}{\partial \bm{o}_{k}} \frac{\partial \bm{o}_{k}}{\partial s_{k}} \right\| \\
&\leq \left( \sigma_{max}^{s} + \sigma_{max}^{a} \right)^{t-j} \label{gradient_upper_bound}
\end{align}
where $\sigma_{max}^{s} = \max_{k} \left\| \frac{\partial \mathcal{T}}{\partial s_{k}} \right\|, \sigma_{max}^{a} = \max_{k} \left\| \frac{\partial \mathcal{T}}{\partial \bm{a}_{k}} \frac{\partial \bm{a}_{k}}{\partial \bm{o}_{k}} \frac{\partial \bm{o}_{k}}{\partial s_{k}} \right\| $.
\end{theorem}
\begin{remark}
The cross-step derivative in Eq.~(\ref{csd_branch}) arises due to the observations being taken from a self-induced state distribution. 
Here we omit latent variables for brevity.
\end{remark}
We first analyze the gradient flow pathways.
Equ.~(\ref{csd_clil}) and Equ.~(\ref{gradient_cl}) indicate that the policy gradient $\frac{\partial J}{\partial \theta}$ can be decomposed into two parts: the first part of $\prod_{k} \frac{\partial \mathcal{T}}{\partial s_{k}}$ that propagates the backward signals through composition of the state transition function; and the second part of $\sum_{n} \left( \frac{\partial \mathcal{T}}{\partial s} \right)^{n} \left( \frac{\partial \mathcal{T}}{\partial \bm{a}} \frac{\partial \bm{a}}{\partial \bm{o}} \frac{\partial \bm{o}}{\partial s}  \right)^{t-j-n} $ that involves passing through the policy multiple times, as illustrated in Fig.~\ref{fig:gradient_clil}. 
Specifically, the first part is described by the product of the Jacobian w.r.t. the state of the agent dynamic function, which is shown in Equ.~(\ref{jacobian_bicycle}) and Equ.~(\ref{jacobian_delta}). Given the relatively small values of the non-diagonal entries, especially at low speed, the Jacobian is numerically close to an identity matrix, i.e., $\frac{\partial \mathcal{T}}{\partial s} \approx I$. 
This indicates that the gradient pathway of $\prod_{k} \frac{\partial \mathcal{T}}{\partial s_{k}}$ can serve as a ``gradient highway" \cite{he2016deep, he2016identity} that facilitates the gradient propagation from future states to any preceding steps.
On the other hand, the gradient pathways $\left( \frac{\partial \mathcal{T}}{\partial \bm{a}} \frac{\partial \bm{a}}{\partial \bm{o}} \frac{\partial \bm{o}}{\partial s}  \right)^{t-j-n} $ propagate between the actions and the observations across different time steps. These pathways inform the policy how the actions made at current step affect the observations visited in the future.
Further, Equ.~(\ref{gradient_agent_interaction}) indicates that this cross-step interaction can be extended among any agents, as shown in Fig.~\ref{fig:gradient_interaction}. For example, the term $\frac{\partial \mathcal{T}}{\partial a^{i}_{t+1}} \frac{\partial a^{i}_{t+1}}{\partial o^{i}_{t+1}} \frac{\partial o^{i}}{\partial s_{t+1}} \frac{\partial \mathcal{T}}{\partial a^{j}_{t}} \frac{\partial a^{j}_{t}}{\partial o^{j}_{t}} \frac{\partial o^{j}}{\partial s_{t}}$ describes how agent $j$ behaves at step $t$ affects agent $i$ at step $t+1$. 
This cross-step inter-agent interaction modelling is facilitated by the gradient highway $\left( \frac{\partial \mathcal{T}}{\partial s} \right)^{n}$ to long temporal intervals.
The existence of gradient highways and cross-step inter-agent gradient pathways confers efficient credit assignment~\cite{minsky1961steps} for closed-loop differentiable simulation, thus ameliorating the causal confusion issue~\cite{de2019causal}.

\subsubsection{Issues}
We then analyze the gradient norm.
Ineq.~(\ref{gradient_upper_bound}) shows that the upper bound of the gradient norm $\left\| \frac{\partial J}{\partial \theta} \right\| $ undergoes exponential changes with increasing time intervals.
Given that the norm of the Jacobian w.r.t. state is typically close to one, i.e., $\left\| \frac{\partial \mathcal{T}}{\partial s} \right\| \approx 1$, we have $\sigma_{max}^{s} + \sigma_{max}^{a} \geq 1 $, which means that the upper bound grows exponentially as time interval $t-j$ increases.
This indicates that policy training via closed-loop differentiable simulation is unstable, prone to experiencing gradient exploding rather than suffering from gradient vanishing.
Note that the gradient exploding issue is determined by the training paradigm, and cannot be fundamentally resolved through gradient clipping or network design.

Another issue roots in forward KL divergence's zero avoiding nature~\cite{murphy2012machine}.
Since optimizing the ELBO is equivalent to maximizing the marginal log-likelihood, which in turn means minimizing the forward KL divergence, the zero avoiding phenomenon thereby occurs when maximizing the ELBO objective.
Specifically, in Equ.~(\ref{ELBO}), when $p_{r}(\tau, \zeta, \xi) \rightarrow 0$ and $p_{\theta}(\tau, \zeta, \xi) > 0$, we have $p(\tau) q_{\phi}(\zeta|\tau) p(\xi|\tau) \log \frac{p_{\theta}(\tau, \zeta, \xi)}{q_{\phi}(\zeta|\tau) p(\xi|\tau)} \rightarrow 0 $ regardless of the value of $p_{\theta}(\tau, \zeta, \xi)$.
Instead, when $p_{r}(\tau, \zeta, \xi) > 0$ and $p_{\theta}(\tau, \zeta, \xi) \rightarrow 0$, we have $p(\tau) q_{\phi}(\zeta|\tau) p(\xi|\tau) \log \frac{p_{\theta}(\tau, \zeta, \xi)}{q_{\phi}(\zeta|\tau) p(\xi|\tau)} \rightarrow -\infty $.
Therefore, while maximizing ELBO encourages the model distribution $p_{\theta}(\tau, \zeta, \xi)$ to cover all trajectory points with $p_{r}(\tau, \zeta, \xi) > 0$, it provides hardly any supervisory signals in regions of low density with $p_{r}(\tau, \zeta, \xi) \approx 0$.
This indicates that the policy is weakly constrained in low density region, and may produce diverse but potentially unrealistic behaviors with a lack of safety awareness, as shown in Fig.~\ref{fig:ablation_framework}.

\subsection{Open-Loop Model-Based Imitation Learning}\label{sec:open-loop}

\subsubsection{Objective}
To stabilize the training of closed-loop model-based IL, we propose the open-loop model-based IL regularization, whose basic idea is to narrow the function space to a range that enables stable training:
\begin{equation}
\begin{aligned}
\max_{\theta, \phi} \quad & \mathbb{E}_{\tau \sim \mathcal{D}, \xi \sim p(\kappa|\tau)} \left[\text{ELBO-CL}(\bm{q}, \tau, \xi;\theta, \phi)  \right] \\
\text{s.t.} \quad & \mathbb{E}_{\tau \sim \mathcal{D}, \xi \sim p(\xi|\tau)} \left[\text{ELBO-OL}(\bm{q}, \tau, \xi;\theta, \phi)  \right] \geq \epsilon
\end{aligned}
\end{equation}
where $\text{ELBO-OL}(\bm{q}, \tau, \xi;\theta, \phi)$ denotes the open-loop model-based IL objective as:
\begin{align}
\label{elbo_ol}
&\mathbb{E}_{\tau \sim \mathcal{D}, \xi \sim p(\xi|\tau)} \left[\text{ELBO-OL}(\bm{q}, \tau, \xi;\theta, \phi)  \right] \nonumber \\
=& \mathbb{E}_{\tau, \xi}  \mathbb{E}_{\zeta \sim \bm{q}_{\phi}(\zeta|\tau), \hat{\tau} \sim p_{\theta}(\hat{\tau}, \zeta, \xi)} [ \sum_{t=0}^{T-1} \log p(s_{t+1}|\hat{s}_{t}, \bm{\pi}_{\theta}( \bm{o}_{\leq t}, \bm{z}_{t}), u_{t})   \nonumber \\
-&  D_{\text{KL}} \left(\bm{q}_{\phi}(\bm{z}_{t}|\tau) \| \bm{\eta}_{\theta}(\bm{z}_{t}|\mathbf{o}_{\leq t}) \right) ].
\end{align}
In this open-loop objective, the policies' observations are from human's state distribution.
However, different from behavior cloning, the analytic environment model is employed to generate the states autoregressively based on the actions output by the policy, thereby overcoming the i.i.d. state-action pairs assumption in BC.
The open-loop objective can be viewed as a necessary condition for the optimal policy. By maximizing this objective, we constrain the function space to a relatively large neighborhood around the optimal policy.
Within this neighborhood, the trajectory generated autoregressively by the policies stays relatively close to the ground-truth trajectory, leading to a loss value that avoids the occurrence of exploding gradients.

\subsubsection{Gradient property}
We then investigate the gradient property of the open-loop model-based objective~(\ref{elbo_ol}) by deriving the following proposition:
\begin{theorem}
Given the assumptions stipulated in Proposition~\ref{thm:general}, the cross-step derivative in open-loop model-based imitation learning is,
\begin{equation}
\label{csd_olil}
\prod_{k} \frac{\partial s_{k+1}}{\partial s_{k}} = \prod_{k} \frac{\partial \mathcal{T}}{\partial s_{k}}
\end{equation}
where $s_{k+1} = \mathcal{T}(s_{k}, \bm{a}_{k})$.
Therefore the policy gradient is,
\begin{equation}
\label{gradient_open_loop}
\frac{\partial J}{\partial \theta} = \sum_{t=1}^{T} \sum_{j=1}^{t} \frac{\partial J_{t}}{\partial s_{t}} \left( \prod_{k=t-1}^{j} \frac{\partial \mathcal{T}}{\partial s_{k}} \right) \frac{\partial s_{j}}{\partial \bm{a}_{j-1}} \frac{\partial \bm{a}_{j-1}}{\partial \theta}. 
\end{equation}
The upper bound of the 2-norm of the cross-step derivative is derived as,
\begin{align}\label{ineq:olil}
\left\| \prod_{k=t-1}^{j} \frac{\partial s_{k+1}}{s_{k}} \right\| &\leq \prod_{k=t-1}^{j} \left\| \frac{\partial \mathcal{T}}{\partial s_{k}} \right\| \leq (\sigma^{s}_{max})^{t-j}
\end{align}
where $\sigma_{max}^{s} = \max_{k} \left\| \frac{\partial \mathcal{T}}{\partial s_{k}} \right\|$.
\end{theorem}
\begin{remark}
The cross-step derivative expression in Eq.~(\ref{csd_olil}) arises because the gradient cannot backpropagate from the action through the state transition function.
Here we omit the latent variables for brevity.
\end{remark}
As shown in Equ.~(\ref{gradient_open_loop}) and Fig.~\ref{fig:gradient_olil}, gradients propagate from future states back to preceding ones, which requires the policy to be aware of the effect of current action to future states. 
Besides, as analyzed in the Sec.~\ref{sec:closed-loop}, the Jacobian w.r.t. state is typically close to identity matrix, i.e., $\frac{\partial \mathcal{T}}{\partial s_{k}} \approx I$ and $\| \frac{\partial \mathcal{T}}{\partial s_{k}} \| \approx 1$.
Combined with the Ineq.~(\ref{ineq:olil}), it is found that gradients of open-loop model-based IL are stable with low possibility of gradient vanishing or gradient exploding.
Compared to behavior cloning with gradient of $\frac{\partial J}{\partial \theta} = \sum_{t=1}^{T} \frac{\partial J_{t}}{\partial s_{t}} \frac{\partial s_{t}}{\partial \bm{a}_{t-1}} \frac{\partial \bm{a}_{t-1}}{\partial \theta} $ shown in Fig.~\ref{fig:gradient_bc}, optimizing the model-based objective~(\ref{elbo_ol}) constrains the function space to a smaller region, which has better capability of stabilizing the training process.

\subsection{Model-Based Reinforcement Learning}\label{sec:RL}

\begin{figure*}[!t]
\centering
\subfloat[]{\includegraphics[]{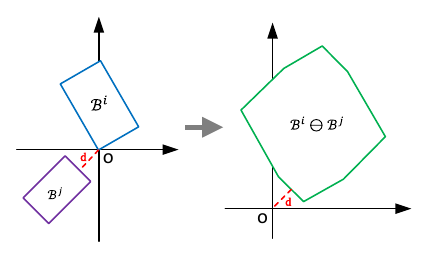}}%
\label{collision_reward}
\hfil
\subfloat[]{\includegraphics[]{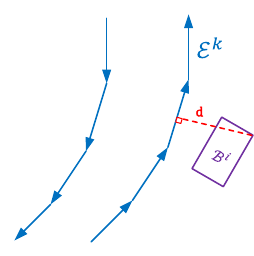}}%
\label{on_road_reward}
\hfil
\subfloat[]{\includegraphics[]{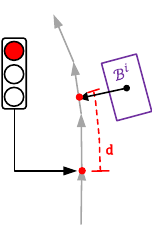}%
\label{traffic_rule_reward}}
\caption{Illustration of differentiable reward functions. (a) Minkowski-difference-based collision avoidance reward. (b) Projection-distance-based on-road reward. (c) Projection-length-based traffic rule compliance reward.}
\label{fig:reward}
\end{figure*}

\subsubsection{Objective}
To constrain the policies' behavior in low-density region, we propose the model-based RL regularization, whose key idea is to leverage explicit reward signals to prevent the policy from having high probability in regions of low real density. 
This is achieved by narrowing the function space to a region with domain knowledge as,
\begin{equation}
\label{overall_objective}
\begin{aligned}
\max_{\theta, \phi} \quad & \mathbb{E}_{\tau \sim \mathcal{D}, \xi \sim p(\xi|\tau)} \left[\text{ELBO-CL}(\bm{q}, \tau, \xi;\theta, \phi)  \right] \\
\text{s.t.} \quad & 
\begin{aligned}[t]
\mathbb{E}_{\tau \sim \mathcal{D}, \xi \sim p(\xi|\tau)} \left[\text{ELBO-OL}(\bm{q}, \tau, \xi;\theta, \phi)  \right] &\geq \epsilon \\
\mathbb{E}_{\tau \sim \mathcal{D}, \xi \sim p(\xi|\tau)} \left[ \mathcal{J}(\hat{\tau};\theta) \right] &\geq \iota
\end{aligned}
\end{aligned}
\end{equation}
where $\mathcal{J}(\hat{\tau};\theta)$ denotes the objective of model-based RL as,
\begin{align}
\label{objective_rl}
&\mathbb{E}_{\tau \sim \mathcal{D}, \xi \sim p(\xi|\tau)} \left[ \mathcal{J}(\hat{\tau};\theta) \right] \nonumber \\
=& \mathbb{E}_{\tau \sim \mathcal{D}, \xi \sim p(\xi|\tau)} \mathbb{E}_{(\hat{\tau},\hat{\zeta}) \sim p_{\theta}(\hat{\tau}, \hat{\zeta}, \xi)} \left[ \sum_{t=1}^{T} \sum_{i=1}^{N} \gamma^{t-1} R^{i}(\hat{s}_{t}) \right] \\
=& \mathbb{E}_{\tau, \xi} \mathbb{E}_{\hat{\tau}, \hat{\zeta}} \left[ \sum_{t=0}^{T-1} \sum_{i=1}^{N} \gamma^{t} R^{i}\left(\mathcal{T}(\hat{s}_{t}, \bm{\pi}_{\theta}(\hat{\bm{o}}_{\leq t}, \hat{\bm{z}}_{t}), u_{t})\right) \right]
\end{align}
where $\hat{\zeta} = \{ \bm{\hat{z}}_{t} \}_{t=0}^{T-1}$ denotes the behavioral latents sampled from the prior distribution.
In this objective, we rollout both the high- and low-level policies to generate trajectory samples that are not included in the dataset, and penalize the samples that violate domain knowledge. 
This provides learning signals in regions that are covered much more extensively than the data distribution.
Besides, by designing reasonable rewards, the samples in the long-tailed region of the data distribution are unaffected, which avoids compromising the policies' performance.
By maintaining an appropriate regularization strength, the infraction rate can be reduced from an artificially high level under only the IL objective to a level close to that of the dataset, as shown in Sec.~\ref{sec:experiment}.
Specifically, we construct a series of differentiable rewards to enhance the differentiable simulator, which enables us to optimize the multi-agent RL objective through analytic policy gradient.
Compared to model-free methods~\cite{schulman2015trust, schulman2017proximal, haarnoja2018soft}, using analytic gradient has much higher efficiency in credit assignment and lower variance.
Compared to the learned models~\cite{hafner2019dream, deisenroth2011pilco}, the near-perfect analytic environment model used here gets rid of compounding model bias and has lower computational complexity.

\subsubsection{Differentiable rewards} \label{rewards}
To inject safety-related knowledge into the policies, we propose the following differentiable reward functions,
\begin{equation}
R^{i} = R^{i}_{\text{collision}} + R^{i}_{\text{on-road}} + R^{i}_{\text{traffic-rule}}.
\end{equation}
As shown in Fig.~\ref{fig:reward}, these rewards are constructed based on efficient vectorized representation of the global state \cite{gao2020vectornet}.
Specifically, the collision avoidance reward aims to guide the agent to maintain a safety distance from each other as,
\begin{equation}
R^{i}_{\text{collision}} = \min(d_{\text{object}}^{i}, \epsilon_{1})
\end{equation}
where $d_{\text{object}}^{i}$ is the distance or negative penetration depth between the bounding boxes of ego agent and nearest neighboring object, and $\epsilon_{1}$ is a pre-defined threshold. We compute $d_{\text{object}}^{i}$ by transforming it into the minimal distance from the Minkowski difference of the two bounding boxes to the origin,
\begin{align}
d_{\text{object}}^{i} =& \min_{j \in \mathcal{N}_{i}} d(\mathcal{B}^{i}, \mathcal{B}^{j}) \\
=& \min_{j \in \mathcal{N}_{i}} 
\begin{cases}
\begin{aligned}
\min_{v_{i} \in \mathcal{B}^{i}, v_{j} \in \mathcal{B}^{j}} \| v_{i} - v_{j} \|_{2}, &~\text{if}~\mathcal{B}^{i} \cap \mathcal{B}^{j} =  \emptyset \\
\min_{t \in \{t|\mathcal{B}^{i} \cap (\mathcal{B}^{j} + t) = \emptyset\}} - \| t \|_{2}, &~\text{otherwise}
\end{aligned}
\end{cases} \\
=& \min_{j \in \mathcal{N}_{i}} [2 \mathbbm{1}_{\{\mathcal{B}^{i} \cap \mathcal{B}^{j} =  \emptyset \}} - 1] \min_{v \in \partial (\mathcal{B}^{i} \ominus \mathcal{B}^{j}) } \| v \|_{2} \label{minkowski_difference}
\end{align}
where $\partial (\mathcal{B}^{i} \ominus \mathcal{B}^{j}) $ denotes the boundary of the Minkowski difference of two bounding boxes constructed by their vertices.
Note that the whole calculation of the distance $d_{\text{object}}^{i}$ is accurate and differentiable.
The on-road reward penalizes the agent for being off road during normal driving as,
\begin{equation}
\label{reward:on-road}
R^{i}_{\text{on-road}} = - \max(d_{\text{edge}}^{i}, \epsilon_{2})
\end{equation}
where $d_{\text{edge}}^{i} $ is the distance to the nearest road edge calculated as follows,
\begin{align}
d_{\text{edge}}^{i} &= (2 \mathbbm{1}_{\text{off-road}} - 1) \min_{\mathcal{E}^{k} \in \mathcal{M}} d(\mathcal{B}^{i}, \mathcal{E}^{k}) \\
&= \left[2 \mathbbm{1}_{\{\bar{g}^{*} \times \bar{v}^{*} < 0\}} - 1 \right] \min_{\mathcal{E}^{k} \in \mathcal{M}} \max_{v \in \mathcal{B}^{i}} \min_{g \in \mathcal{E}^{k}} \| \bar{v}- \mu \bar{g} \|_{2}
\end{align}
where $\mathcal{E}^{k} $ denotes the ordered vector sequence of a road edge; $\bar{v} $ and $\bar{g}$ are in the local coordinate of the edge vector $g $, and $\mu \bar{g} $ denotes the clamped projected vector with $\mu \in [0, 1] $. Note that the on-road reward is masked out in situations such as parking or exiting the road from driveway.
Finally, the traffic-rule compliance reward encourages the agent to follow the traffic light. Concretely, we penalize the agent for running a red light as follows,
\begin{equation}
R^{i}_{\text{traffic-rule}} = - \max(\min(d_{\text{light}}^{i}, \epsilon_{3}), \epsilon_{4})
\end{equation}
where $d_{\text{light}}^{i}$ is the projected distance on the lane center beyond the stop point of the red light,
\begin{equation}
d_{\text{light}}^{i} = \text{proj}_{L}(p^{i}) - \text{proj}_{L}(p_{\text{light}})
\end{equation}
where $\text{proj}_{L}(p_{\text{light}})$ and $\text{proj}_{L}(p^{i})$ are the projected length of the red light and the agent on the route, respectively.
Here the vehicle's route is obtained by performing a depth-first search on the road graph. 
Note that the on-road and traffic rule compliance rewards only apply to the vehicle agents.

\subsection{Dynamic Lagrange Multipliers}\label{MTO}
In the constrained optimization problem (COP~\ref{overall_objective}), the closed-loop model-based IL objective provides primary learning signals while the open-loop model-based IL regularization guarantees stable training and the model-based RL regularization constrains the policy in low-density region.

\subsubsection{Optimization challenge}
This problem can be approximately solved by Taylor expansion with conjugate gradient (e.g. TRPO \cite{schulman2015trust}).
Alternatively, we can derive the Lagrangian of this COP as follows,
\begin{align}
\label{unconstrained_objective}
\mathcal{L}(\theta, \phi, \lambda) =& \lambda_{0} \mathbb{E}_{\tau \sim \mathcal{D}, \xi \sim p(\xi|\tau)} \left[  \text{ELBO-CL}(\bm{q}, \tau, \xi;\theta, \phi) \right] \nonumber  \\ 
+& \lambda_{1} \mathbb{E}_{\tau \sim \mathcal{D}, \xi \sim p(\xi|\tau)} \left[  \text{ELBO-OL}(\bm{q}, \tau, \xi;\theta, \phi) \right] \nonumber \\
+& \lambda_{2} \mathbb{E}_{\tau \sim \mathcal{D}, \xi \sim p(\xi|\tau)} \left[ \mathcal{J}(\hat{\tau};\theta)  \right]
\end{align}
where $\lambda_{0} = 1 $ and $\lambda_{1,2}$ are the Lagrange multipliers. 
Then the original problem is equivalent to a min-max optimization problem: $\min_{\theta, \phi} \max_{\lambda} -\mathcal{L}(\theta, \phi, \lambda)$, which can be solved by alternating the inner maximization and outer minimization (e.g. dual gradient descent \cite{boyd2004convex}).
However, both methods have high computational complexity for analytic-gradient-based policy learning.
We chose to simplify the problem by relaxing the hard constrains into soft penalties. 
Nevertheless, using fixed multipliers is insufficient to obtain the optimal policy because the data distribution is constantly changing in the case of closed-loop policy learning.
Excessively large multipliers result in a high condition number of the Hessian matrix, making optimization challenging; while overly small multipliers weaken regularization, leading to degraded policy performance.
Instead of alleviating this issue by adjusting the coefficients heuristically (e.g. PPO-Penalty \cite{schulman2017proximal}), here we proposed a principled method.

\subsubsection{Multiplier equation}
Our key insight is that the optimal multipliers should eliminate the directional conflicts and amplitude dominance of all the gradient components, such that both the objective and the regularizations are guaranteed to be effective at each iteration.
Specifically, we first apply orthogonal approximation to the original gradient matrix $\hat{G}$, 
\begin{equation}
G^{*} = \arg\min_{G}~\left\| G - \hat{G}  \right\|_{\text{F}} \quad \text{s.t.}~G^{T}G = I.
\end{equation}
The above COP is a kind of orthogonal Procrustes problem \cite{schonemann1966generalized} and has closed-form solution of $G^{*} = U V^{T}$, where $\hat{G} = U \Sigma V^{T}$ is the singular value decomposition (SVD) of the original gradient matrix.
The solution $G^{*} $ has two nice properties originating from its orthogonality: (\romannumeral 1) the modified gradients are orthogonal to each other, which eliminates the negative interference of the regularizations with the objective; (\romannumeral 2) the norms of all gradients are equal, which guarantees the effectiveness of the regularizations.
We then establish the following equation to solve for the multipliers,
\begin{equation}
\label{multiplier_equation}
\hat{G} \lambda^{*} = \sigma G^{*} \omega
\end{equation}
where $\lambda = [\lambda_{0}, \lambda_{1}, \lambda_{2}]^{\text{T}}$ is the augmented multiplier vector; $\sigma$ is a scale factor (e.g., mean of singular values) to maintain consistency in gradient matrix's norm before and after approximation, i.e., $\| \hat{G} \| =\| \sigma G^{*} \|$; and $\omega$ is a fixed weight vector that controls the strength of regularizations. 
Since the gradient matrix $\hat{G}$ is typically full rank, the above equation has a unique solution.
Therefore the expression for the multipliers is derived as follows:
\begin{equation}
\lambda^{*} = \sigma V \Sigma^{-1} V^{T} \omega
\end{equation}
with total computational complexity of $\mathcal{O}(\max(|\theta|, |\phi|))$.
The idea of orthogonal gradient is previously applied to restricting the direction of gradient update in continual learning~\cite{farajtabar2020orthogonal}, and reducing the cosine similarity~\cite{suteu2019regularizing} or condition number~\cite{senushkin2023independent} of task gradients in supervised multi-task learning.
However, by dynamically solving for the optimal multipliers in orthogonal sense through the proposed multiplier equation, our method is particularly suitable for constrained policy optimization with continuously changing data distributions. 

\begin{figure*}[!t]
\centering
\includegraphics{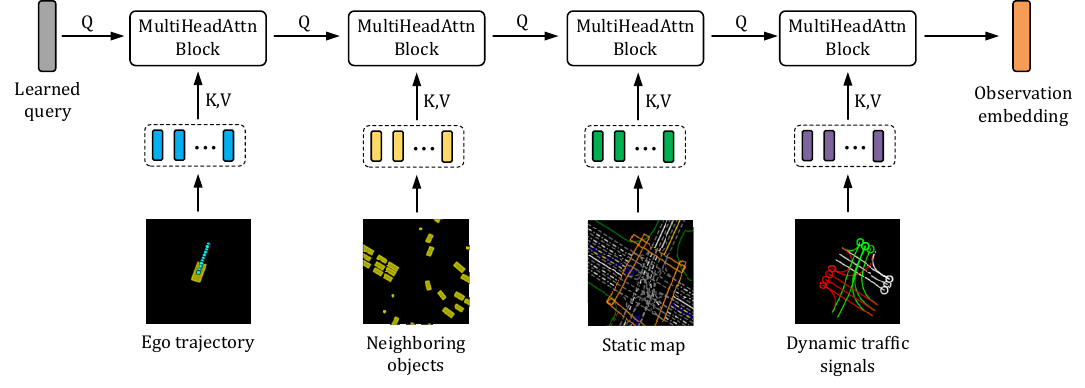}
\caption{Structure of TokenNet.}
\label{TokenNet}
\end{figure*}

\subsection{Function Approximation}\label{sec:net}
Modelling the multi-agent policy is a high-dimensional problem considering the large number of road users and the complex scene contexts in long time horizon.
Therefore, we adopt neural networks with efficient architecture to serve as the function approximators.
To simplify the architecture, we make the following agent-wise factorizations and conditional independence assumptions, 
\begin{align}
\bm{\eta}_{\theta}(\bm{z}_{t} | \bm{o}_{\leq t}) =& \prod_{i=1}^{N} \eta_{\theta}^{i} (z_{t}^{i} | o_{t}^{i}), \\ 
\bm{\pi}_{\theta}(\bm{a}_{t} | \bm{o}_{\leq t}, \bm{z}_{t}) =& \prod_{i=1}^{N} \pi_{\theta}^{i}(a_{t}^{i} | o_{\leq t}^{i}, z_{t}^{i}), \\
\bm{q}_{\phi}(\zeta | \tau) =& \prod_{i=1}^{N} \prod_{t=1}^{T} q^{i}_{\phi} (z_{t}^{i} | \tau^{i}).
\end{align}
As discussed in Prop.~(\ref{thm:clil}), with a differentiable simulator, an agent-wise factored policy can operate in a way similar to centralized training with decentralized execution \cite{lowe2017multi}.

\subsubsection{TokenNet}
A building block of all the following networks is the TokenNet, which converts a high-dimensional observation into a low-dimensional embedding vector as follows,
\begin{equation}
x^{i}_{t} = \phi_{\text{tkn}}(o_{t}^{i} = \{ \mathcal{P}_{\text{ego}}^{i}, \mathcal{P}_{\text{obj}}^{i}, \mathcal{P}_{\text{map}}^{i}, \mathcal{P}_{\text{sig}}^{i} \}),
\end{equation}
where $x^{i}$ is the observation embedding. The tokenize function $\phi_{\text{tkn}}$ iteratively extracts information from each modality of the observation as:
\begin{equation}
x^{i}_{t} = \text{MultiHeadAttn}(x^{i}_{t}, g_{\text{enc}}(\mathcal{P}_{j}^{i}), g_{\text{enc}}(\mathcal{P}_{j}^{i})),
\end{equation}
where MultiHeadAttn() denotes the multi-head attention layer \cite{vaswani2017attention}; $g_{\text{enc}}$ is a multi-layer perceptron (MLP) whose weights are different for each modality; $j$ is the index of each modality. The initial value of $x^{i}_{t}$ is a learnable query.
Leveraging the asymmetry of the cross-attention operator, TokenNet has a linear complexity of $\mathcal{O}(M)$ w.r.t. number of elements in each modality.

\subsubsection{Policy network}
Based on the TokenNet, the high-level policy outputs the index of the embedding vector in the codebook as:
\begin{equation}
k^{i}_{t} \sim \varphi_{\text{idx}}(\phi_{\text{tkn}}(o_{t}^{i})),
\end{equation}
where $k^{i}_{t}$ is a one-hot vector, and $\phi_{\text{idx}}$ is implemented as MLPs.
To obtain the latent variable, we have:
\begin{equation}
z^{i}_{t} = \varphi_{\text{codebook}}(k^{i}_{t}),
\end{equation}
where $\varphi_{\text{codebook}}$ are the linear projections.
For the low-level policy, we design the network structure as follows:
\begin{align}
a_{t}^{i} =& \varphi_{\text{act}} \circ \varphi_{\text{rel}}  \left( \varphi_{\text{temp}}\left( \left\{ \phi_{\text{tkn}} (o_{l}^{i}) \right\}_{l=0}^{t} \right), z_{t}^{i} \right),
\end{align}
where $\varphi_{\text{temp}}$ is the temporal aggregation operator implemented as the gated recurrent unit \cite{chung2014empirical}; $\varphi_{\text{rel}}$ is relational operator implemented as concatenation; $\varphi_{\text{act}}$ are implemented as MLPs.
Note that the weights of $\varphi_{\text{codebook}}$ and $\varphi_{\text{act}}$ are not shared among different types of agents, and $\varphi_{\text{tkn}}$ are different for high- and low-level policies.

\subsubsection{Posterior network} 
Posterior network infers the behavioral latents using the logged sequence of state-observation pairs as:
\begin{equation}
\zeta^{i} \sim \varphi_{\text{idx}} \circ \varphi_{\text{agg}} \circ \varphi_{\text{inter}} \left( \left\{ \varphi_{\text{rel}} \left(\phi_{\text{tkn}}(o_{l}^{i}), g_{\text{enc}}(s_{l}^{i}) \right)\right\}_{l=0}^{T} \right)
\end{equation}
where the states $s_{l}^{i}$ are normalized w.r.t. the initial pose of the agent; $\varphi_{\text{rel}}$ is a relational operator implemented as the element-wise addition; $\varphi_{\text{agg}}$ is the maxpooling operation used to aggregate information within a time interval. The interaction operator $\varphi_{\text{inter}}$ captures the temporal relationship between the state-observation pairs via self-attention mechanism,
\begin{equation}
Y^{i} = \text{MultiHeadAttn}(Y^{i}, Y^{i}, Y^{i})
\end{equation}
where $y_{t}^{i}$ is the embedding vector of the state-observation pair.

\section{Experiment and Results}\label{sec:experiment}

\subsection{Experimental Setup}

\subsubsection{Dataset}
We conducted the experiments using the large scale Waymo Open Motion Dataset (WOMD \cite{ettinger2021large}).
WOMD contains more than $570$ hours of data covering over $1750$ km of roadways and including more than $100,000$ diverse driving scenes.
Each scene consists of bounding box tracks of highly interactive agents of different types (e.g. vehicle, pedestrian and cyclist), traffic signal states sampled at $10\mathrm{Hz}$, and corresponding high definition vectorized map.
These data are further processed into $487K$ training and $49K$ validation segments of $9\mathrm{s}$ length. 
Each segment provides labels for up to $8$ agents selected to encompass all agent types and exhibit non-trivial behaviors.
In this work, we used the above training and validation splits to train and test the proposed method respectively.
The segments that contain more than $256$ agents and $10K$ lane center points were excluded due to memory constraints, resulting in $428K$ training and $39K$ testing segments.
$450$ segments in the training splits were used for validation to select the training checkpoint for evaluation.

\subsubsection{Simulation setup}
Each simulation lasted for $9\mathrm{s}$ with frequency of $5\mathrm{Hz}$.
The first $1\mathrm{s}$ of each segment were used to initialize the simulation while the remaining $8\mathrm{s}$ were used for training or evaluation.
During training, the labeled agents in the segment were controlled by the learned policy while other agents were replayed from the log. 
During inference, all active agents, such as non-parked vehicles, were controlled by the policy to account for potential interactions with the SDV in the simulation task.
Evaluation metrics were calculated using the simulated dynamic states of the labeled agents, given the integrity of their logged states and the limited size of the context map.

\subsubsection{Implementation details}
We provide the details and hyperparameters of observation design, network architecture, simulation setup, training process, constructing histograms, and search algorithm in the appendix.

\subsection{Metrics}
We employed a series of metrics to evaluate the proposed framework from three aspects, including scenario reconstruction, behavioral realism, and distributional matching.

\subsubsection{Scenario reconstruction}
An accurate behavior model should be able to generate diverse and realistic trajectories that cover the logged scenarios in terms of spatial distance. Therefore, 
we used minADE, minSADE, and ADE to evaluate different models from both agent- and scene-level. Specifically, 
minADE measures the diversity of the sampled trajectory at the agent-level as follows,
\begin{equation*}
\text{minADE} =  \frac{1}{N_{e}T} \sum_{i=1}^{N_{e}} \min_{k} \sum_{t=1}^{T} \| \hat{s}_{t}^{i,k} - s_{t}^{i} \|_{2}
\end{equation*}
minSADE measures the diversity of the sampled scenes as:
\begin{equation*}
\text{minSADE} = \min_{k} \frac{1}{N_{e}T} \sum_{i=1}^{N_{e}} \sum_{t=1}^{T} \| \hat{s}_{t}^{i,k} - s_{t}^{i} \|_{2}
\end{equation*}
ADE measures the precision of the samples as:
\begin{equation*}
\text{ADE} = \frac{1}{K_{r}N_{e}T} \sum_{k=1}^{K_{r}} \sum_{i=1}^{N_{e}} \sum_{t=1}^{T} \| \hat{s}_{t}^{i,k} - s_{t}^{i} \|_{2}
\end{equation*}
where $K_{r} = 16$ is the number of simulation rollouts and $N_{e}$ is the number of agents evaluated.

\subsubsection{Behavior realism}
We measured the realism of the generated behaviors using the following four metrics from the perspectives of agent-agent interaction, agent-map interaction and kinematics. Concretely, collision rate evaluates the consistency of agents' behaviors as follows:
\begin{equation*}
\text{Collision Rate} = \frac{1}{K_{r} N_{e}} \sum_{k=1}^{K_{r}} \sum_{i=1}^{N_{e}} \min \left(1, \sum_{t=1}^{T} \mathbbm{1} \left[ d_{\text{object}}^{i,t,k} < 0 \right] \right)
\end{equation*}
Off-road rate calculates the percentage of vehicle agents that drive off roads without passing through the driveways as,
\begin{equation*}
\text{Off-Road Rate} = \frac{1}{K_{r} N_{e}} \sum_{k=1}^{K_{r}} \sum_{i=1}^{N_{e}} \min \left(1, \sum_{t=1}^{T} \mathbbm{1} \left[ d_{\text{edge}}^{i,t,k} > 0 \right] \right)
\end{equation*}
Traffic rule violation rate computes the percentage of vehicle agents that run a red light as,
\begin{equation*}
\text{Rule Vio. Rate} = \frac{1}{K_{r} N_{e}} \sum_{k=1}^{K_{r}} \sum_{i=1}^{N_{e}} \min \left(1, \sum_{t=1}^{T} \mathbbm{1} \left[ d_{\text{light}}^{i,t,k} > 0 \right] \right)
\end{equation*}
where $d_{\text{object}}$, $d_{\text{edge}}$ and $d_{\text{light}}$ are described in Sec.~\ref{rewards}.
Kinematics infeasibility rate calculates the percentage of vehicle agents whose transitions are kinematically infeasible using estimated acceleration and steering curvature from the states,
\begin{align*}
&\text{Kinematic Infeasibility Rate} \\
=&\frac{1}{K_{r} N_{e}} \sum_{k=1}^{K_{r}} \sum_{i=1}^{N_{e}} \min \left(1, \sum_{t=1}^{T} \mathbbm{1} \left[ |a_{t}^{i,k}| > \bar{a} \right] + \mathbbm{1} \left[ |\kappa_{t}^{i,k}| > \bar{\kappa} \right] \right)
\end{align*}
where $\bar{a} = 6\mathrm{m/s^{2}}$ and $\bar{\kappa} = 0.3\mathrm{m^{-1}}$ are the maximum acceleration and curvature respectively.

\subsubsection{Distribution divergence}
We evaluated distributional divergence between the learned policy and human demonstrations for both state visitation distribution $\nu_{\pi}(s)$ and trajectory distribution $p_{\pi}(\tau)$. 
To address the curse of dimensionality, we extracted scalar features from state and trajectory. 
Three groups of state features were considered, including linear and angular speed and acceleration (kinematics), distance to nearest object and time-to-collision (agent-agent interaction), and distance to road edge (agent-map interaction). 
Trajectory features include average curvature to measure behavioral intent and progress to evaluate behavioral style.
The distributions over these features were approximated by histograms. 
Then we calculated Jensen-Shannon divergence (JSD) on these histograms as the distribution matching metric.

\subsection{Comparison with Baselines}\label{sec:main_results}

\begin{table*}[ht]
\centering
\caption{Comparison with baselines on Waymo Open Motion Dataset. The non-zero collision rate of the dataset is because the labelled bounding boxes are slightly larger than the true size of agents, especially for pedestrians. The speed JSD refers to the average of linear and angular speed JSD. The same applies to acceleration JSD.}
\begin{tabular}{*{15}{c}}
\toprule
 & \multicolumn{3}{c}{Reconstruction (m) $\downarrow$} & \multicolumn{4}{c}{Behavior realism (\%) $\downarrow$} & \multicolumn{7}{c}{Distribution matching JSD ($\times {10}^{-3}$) $\downarrow$} \\
\cmidrule(r){2-4} \cmidrule(lr){5-8} \cmidrule(lr){9-15}
Methods & \makecell[c]{Min \\ ADE} & \makecell[c]{Min \\ SADE} & ADE & Coll. & \makecell[c]{Off- \\ road} & \makecell[c]{Rule \\ vio.} & \makecell[c]{Kin. \\ inf.} & Spd. & Acc. & \makecell[c]{Dist.2 \\ obj.} & TTC & \makecell[c]{Dist.2 \\ edge} & Curv. & Prog. \\
\midrule
Log & - & - & - & 3.34 & 1.53 & 3.53 & 1.40 & - & - & - & - & - & - & - \\
\cmidrule(lr){1-15}
SimNet~\cite{bergamini2021simnet} & 4.57 & 4.57 & 4.57 & 14.22 & 11.11 & 4.91 & 0.01 & 5.27 & 76.00 & 8.92 & 19.10 & 4.54 & 2.72 & 6.33 \\
UrbanDriver~\cite{scheel2022urban} & 3.20 & 3.20 & 3.20 & 8.35 & 10.51 & 4.12 & 30.79 & 32.45 & 98.18 & 5.54 & 28.03 & 9.21 & 59.57 & 1.60 \\
RTR~\cite{zhang2023learning} & 3.01 & 3.07 & \underline{3.15} & 8.91 & 4.38 & 4.46 & 0.03 & 7.29 & 47.77 & 3.95 & 4.66 & \textbf{0.98} & 6.98 & 1.24 \\
VAE-MGAIL~\cite{wang2017robust} & 3.98 & 4.62 & 5.57 & 28.74 & 16.12 & 6.68 & \textbf{0.} & 3.02 & 35.78 & 15.53 & 68.78 & 6.32 & 6.86 & 2.20 \\
RTC~\cite{igl2023hierarchical} & 1.60 & 4.48 & 9.45 & 35.32 & 4.81 & 9.83 & 5.22 & 8.61 & 9.81 & 23.53 & 85.88 & 1.23 & 12.19 & 20.56 \\
ITRA~\cite{scibior2021imagining} & 1.74 & 2.95 & 5.08 & 18.38 & 5.00 & 7.56 & \textbf{0.} & 10.46 & 13.51 & 12.98 & 34.58 & 1.34 & 10.06 & 8.40 \\
TrafficSim~\cite{suo2021trafficsim} & 2.62 & 3.36 & 4.53 & 14.53 & 4.69 & 3.62 & 0.13 & 31.49 & 47.33 & 8.94 & 17.13 & 1.98 & 4.62 & 15.02 \\
MRIC (Ours) & \textbf{1.15} & \textbf{1.97} & \textbf{3.68} & \textbf{2.56} & \textbf{2.56} & \textbf{3.26} & \textbf{0.} & \textbf{1.92} & \textbf{5.81} & \textbf{3.50} & \textbf{1.51} & 4.56 & \textbf{1.30} & \textbf{1.21} \\
\bottomrule
\end{tabular}
\label{tab:main_result}
\end{table*}

\begin{figure}[!t]
\centering
\includegraphics{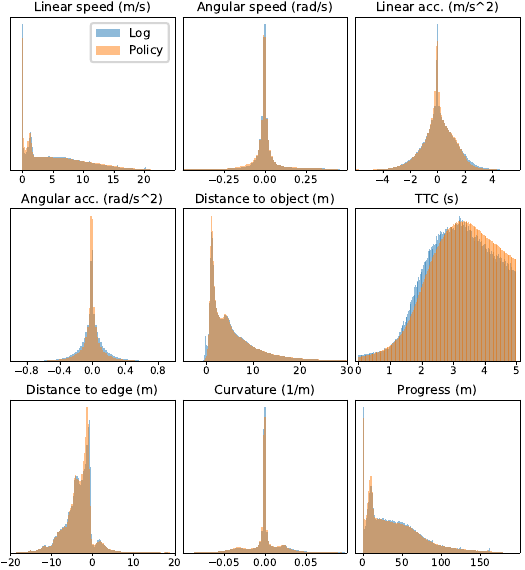}
\caption{Distribution comparison of state features and trajectory features between \texttt{MRIC} and dataset.}
\label{fig:histograms}
\end{figure}

\subsubsection{Baselines}
We re-implemented a series of recent works to compare them with the proposed method.
To make fair comparisons, we extended these works to model the behaviors of different types of agents and adopted the same network architecture described in Sec.~\ref{sec:net} for all baselines.
Specifically, 
\texttt{SimNet}~\cite{bergamini2021simnet} is an open-loop motion prediction model that outputs deterministic future trajectory. \texttt{UrbanDriver}~\cite{scheel2022urban} trained the deterministic policy in closed-loop using a differentiable simulator. \texttt{RTR}~\cite{zhang2023learning} further enhanced this approach by combining it with the model-free PPO~\cite{schulman2017proximal} algorithm, using a stochastic policy. 
These three works focus on mimicking unimodal behaviors, while the following baselines further consider multi-modality.
\texttt{VAE-MGAIL} is a model-based variant of VAE-GAIL~\cite{wang2017robust} that combined an open-loop VAE objective with a closed-loop GAIL objective.
\texttt{RTC}~\cite{igl2023hierarchical} performed posterior training with VAE and GAIL objectives, and prior training using GAIL objective.
Both \texttt{TrafficSim}~\cite{suo2021trafficsim} and \texttt{ITRA}~\cite{scibior2021imagining} trained their policies in closed-loop using the VAE objective. While \texttt{ITRA} employed a bicycle model, \texttt{TrafficSim} predicted future trajectories and implemented collision penalties on the prior samples.
We added the open-loop model-based IL objective from Sec.~\ref{sec:open-loop} to all baselines, except for \texttt{SimNet} and \texttt{TrafficSim}, since it greatly stabilizes training.
We also employed spectral normalization~\cite{miyato2018spectral} for \texttt{VAE-MGAIL} and \texttt{RTC} to improve training stability.

\subsubsection{Results}
Table.~\ref{tab:main_result} shows the quantitative comparison results between MRIC and baselines.
The pure open-loop method \texttt{SimNet} performs poorly due to the covariate shift~\cite{ross2011reduction} and causal confusion~\cite{de2019causal} issues.
The introduction of closed-loop training via differentiable simulation in \texttt{UrbanDriver} significantly improves the performance (e.g., ADE and collision rate).
Combining the PPO objective in \texttt{RTR} can further enhance realism (e.g., off-road rate and TTC JSD), though its effectiveness remains limited (e.g., collision rate and acceleration JSD).
Furthermore, these unimodal methods are unable to capture the diversity of human behaviors (e.g., minADE and minSADE).
For multi-modal baselines, \texttt{VAE-MGAIL}, which relies solely on the GAIL objective in closed-loop training, exhibited low performance. We argue that the discriminator representation used for classification can hardly provide meaningful learning signals for the challenging multi-agent behavior simulation task~\cite{huang2023policy}.
Although \texttt{RTC} further combined VAE objective and prior training, its performance is still unsatisfactory (e.g., collision rate).
Instead, \texttt{ITRA} and \texttt{TrafficSim}, which only employ VAE objective, obtain better multi-modal performance (e.g., minSADE and collision rate).
Compared to the adversarial objective, the learning signals from the state-matching objective are found to be much more effective for our task.
It also appears that the effect of collision penalty, implemented on prior samples in \texttt{TrafficSim}, is marginal.

Finally, our proposed \texttt{MRIC} framework performed the best across most metrics.
In scenario reconstruction, \texttt{MRIC} outperformed other baselines by large margins in terms of both diversity (minADE and minSADE) and precision (ADE). This is attributed to the proposed mixture-of-codebooks module with a temporal abstraction mechanism, which effectively models the uncertainty and variation scale of human behaviors.
In behavioral realism, the combination of model-based RL regularization and a dynamic multipliers mechanism significantly reduces the collision rate, off-road rate, and rule violation rate. Furthermore, the use of a near-perfect dynamic model guarantees kinematic feasibility. These factors lead to realistic behaviors generated by \texttt{MRIC}.
In distribution matching, \texttt{MRIC} achieved the lowest overall JSD for both state features (e.g., acceleration and TTC) and trajectory features (e.g., curvature and progress). This indicates that both state-visitation and trajectory distribution induced by \texttt{MRIC} closely match the data distribution. The distributional realism of \texttt{MRIC} is ascribed to the effective integration of model-based IL and RL via differentiable simulation.
We also note a trade-off between the off-road rate and the distance to road edge JSD, due to the selection of a relatively large threshold (i.e., $-1\mathrm{m}$) for the on-road reward. This trade-off can be better balanced by fine-grained hyper-parameter search as needed. Fig.~\ref{fig:histograms} further presents the histograms of state and trajectory features for \texttt{MRIC} and the dataset, demonstrating that both the state-visitation and trajectory distributions induced by \texttt{MRIC} are very naturalistic and human-like.

\subsection{Behavior Analysis}
In the form of case studies, we further analyze the advantages of the proposed method, including the ability to generate multimodal behaviors in highly interactive scenarios, the adaptability to diverse road topologies, and the capability to precisely simulate irregular human behaviors.

\subsubsection{Multi-modal behaviors}
\begin{figure*}[!t]
\centering
\captionsetup[subfigure]{labelformat=empty} 
\subfloat[]{\includegraphics[]{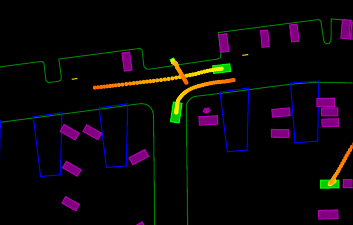}}\label{}\hfil
\subfloat[]{\includegraphics[]{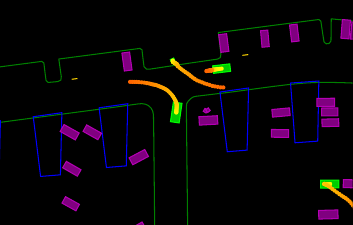}}\label{}\hfil
\subfloat[]{\includegraphics[]{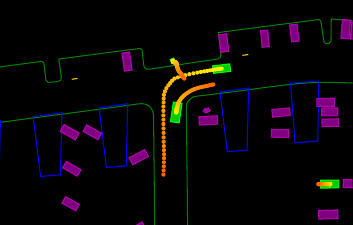}}\label{}\\
\vspace{-0.7cm}
\subfloat[]{\includegraphics[]{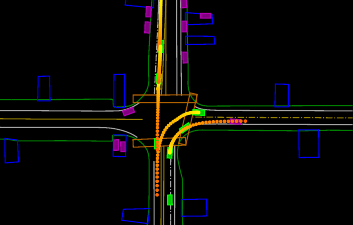}}\label{}\hfil
\subfloat[]{\includegraphics[]{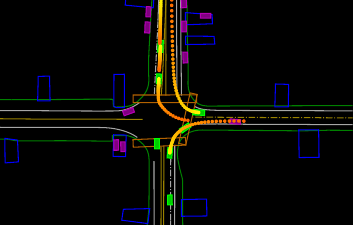}}\label{}\hfil
\subfloat[]{\includegraphics[]{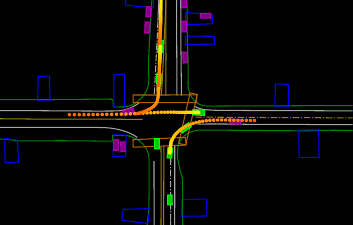}}\label{}\\
\vspace{-0.7cm}
\subfloat[]{\includegraphics[]{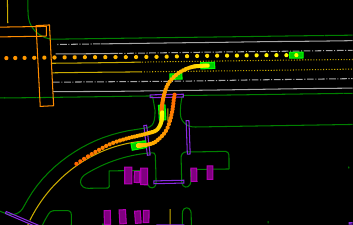}}\label{}\hfil
\subfloat[]{\includegraphics[]{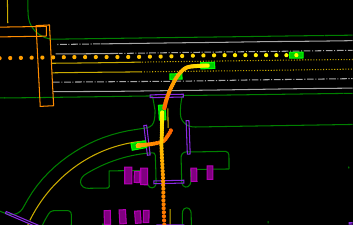}}\label{}\hfil
\subfloat[]{\includegraphics[]{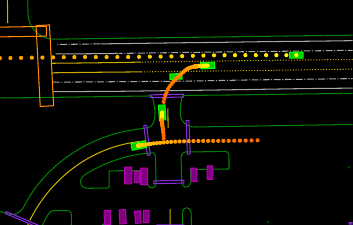}}\label{}\\
\vspace{-0.4cm}
\caption{Illustration of multi-modal behaviors generated by \texttt{MRIC}. Color gradient from yellow to orange indicates time progression. Lime green and fuchsia agents were controlled by the policy and replayed, respectively. Green, blue, purple, and orange polylines indicate road edges, driveways, speed bumps, and crosswalks, respectively. The following is the same. Dynamic visualizations: 
\href{https://drive.google.com/file/d/1gliYK0sh8AB2zmU1orbpXVSojK9BGrL8/view?usp=sharing}{Scenario-1 Mode-1},
\href{https://drive.google.com/file/d/1I-6aJgGwdaA1JC6KtEzD0bzaQjqZwgxy/view?usp=sharing}{S1M2},
\href{https://drive.google.com/file/d/11soezkDFk3wNSexuXszMhvpA9n_3TSTU/view?usp=drive_link}{S1M3};
\href{https://drive.google.com/file/d/13GrdqAFCr8Dx3zDIIUtHmrpozj1DKmAp/view?usp=sharing}{S2M1},
\href{https://drive.google.com/file/d/1yuyRZmoIz2Gse3miNnJW4Tx6gcxwiCzE/view?usp=sharing}{S2M2},
\href{https://drive.google.com/file/d/1f5E39QEx_E0CPfkZY3K_ZhXPqfHIi1M8/view?usp=sharing}{S2M3};
\href{https://drive.google.com/file/d/1BLcrC4py9z1eTBpe3T_mlp-p5K9JEoAW/view?usp=sharing}{S3M1},
\href{https://drive.google.com/file/d/1n85WxHwwqmmtcuon3bOf7a_ZKy_kc21W/view?usp=sharing}{S3M2},
\href{https://drive.google.com/file/d/1CqONCnF1tGXQwCaY7brWpawNCJy_DGZB/view?usp=sharing}{S3M3}.
}
\label{fig:multi-modal}
\end{figure*}

Fig.~\ref{fig:multi-modal} shows the future rollouts simulated by \texttt{MRIC} in three interactive scenarios, where dynamic visualizations are also provided.
In scenario 1, a pedestrian crosses the road either before or after two neighboring vehicles pass, while the vehicles may proceed straight or turn left or right.
In scenario 2, the vehicle arriving the junction first may go straight or make a left or right turn, while another vehicle arriving later yields to this vehicle and proceeds with similar possibilities.
In scenario 3, a vehicle either leaves the parking lot at different times according to the intent of the entering vehicle, or goes straight resisting the entering vehicle.
These examples illustrate the ability of \texttt{MRIC} to capture distinct modes of the trajectory distribution, thereby producing diverse and realistic multi-agent behaviors.

\subsubsection{Map-adaptive behaviors}
\begin{figure*}[!t]
\centering
\captionsetup[subfigure]{labelformat=empty} 
\hfil\subfloat[]{\includegraphics[]{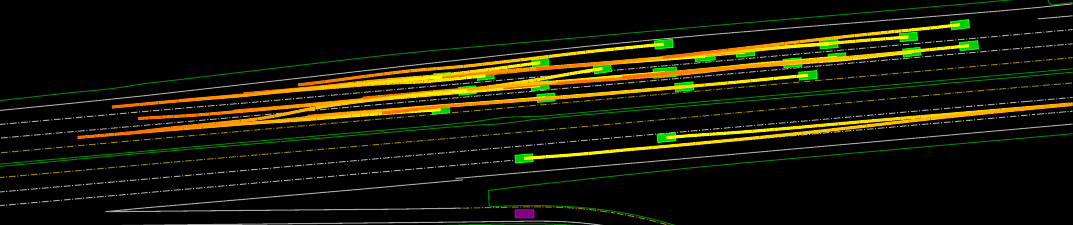}}\label{}\hfil\\
\vspace{-0.7cm}
\subfloat[]{\includegraphics[]{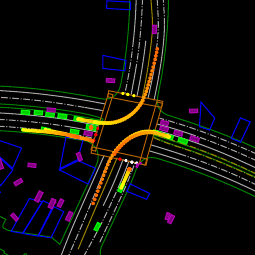}}\label{}\hfil
\subfloat[]{\includegraphics[]{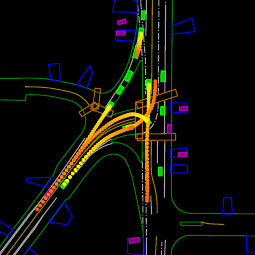}}\label{}\hfil
\subfloat[]{\includegraphics[]{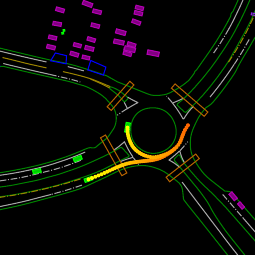}}\label{}\hfil
\subfloat[]{\includegraphics[]{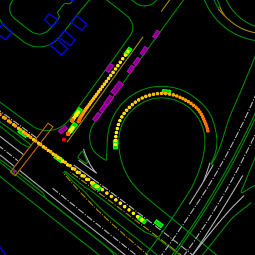}}\label{}\hfil\\
\vspace{-0.4cm}
\caption{Illustration of map-adaptive behaviors produced by \texttt{MRIC}. Dynamic visualizations:
\href{https://drive.google.com/file/d/1DhZi5CLJttcE7p6QBLt9Mftz-RKX05rq/view?usp=sharing}{highway},
\href{https://drive.google.com/file/d/1sGwwxCTsKvpW2c7D1LqKTLEI1_Z8Tiaq/view?usp=sharing}{signalized} and
\href{https://drive.google.com/file/d/1NS2gFaRm6RGciNkRPhQp-G7tjgdLFXT4/view?usp=sharing}{unsignalized intersections},
\href{https://drive.google.com/file/d/1UeiYoNLo5Vc0Nb6XoEBOdn_I0zla-Lka/view?usp=sharing}{roundabout} and
\href{https://drive.google.com/file/d/1jjHVpq3DBJipJbMqf0aKfVfiJS6LAPR6/view?usp=sharing}{ramp}.
}
\label{fig:map-adaptive}
\end{figure*}

Fig.~\ref{fig:map-adaptive} shows the behaviors generated by \texttt{MRIC} in typical traffic scenarios with diverse road topologies. 
On the highway, a large number of vehicles are simultaneously controlled, exhibiting both lane-keeping and lane-changing behaviors. These vehicles maintained a safe distance between each other, similar to human drivers, in long-distance simulations.
At the signalized intersection, vehicles turning left followed the green light to pass through, while the vehicle and cyclist turning right stopped and yielded to obey the red light.
At the unsignalized intersection, vehicles from different directions passed through alternately; in merges, vehicles adjusted their speed to facilitate traffic merging safely.
In the roundabout, the circulating vehicle smoothly traversed the central island, while the entering vehicle decelerated and yielded to it.
On the entry ramp, two controlled vehicles smoothly followed the curve with gentle steering angles.
Fig.~\ref{fig:multi-modal} also shows the scenarios of T-junction and parking lot.
These case studies demonstrate the applicability of \texttt{MRIC} to various traffic scenarios with distinct context maps.

\subsubsection{Irregular behaviors}
\begin{figure*}[!t]
\centering
\captionsetup[subfigure]{labelformat=empty} 
\subfloat[]{\includegraphics[]{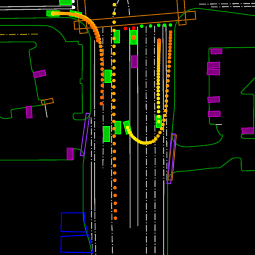}}\label{}\hfil
\subfloat[]{\includegraphics[]{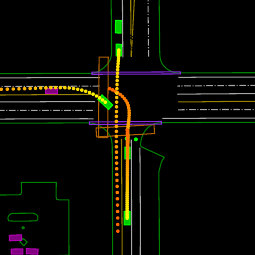}}\label{}\hfil
\subfloat[]{\includegraphics[]{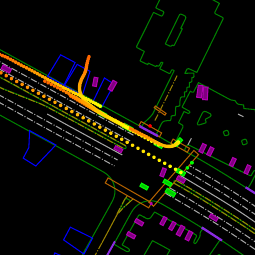}}\label{}\hfil
\subfloat[]{\includegraphics[]{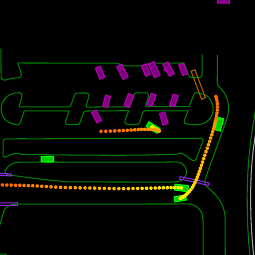}}\label{}\hfil\\
\vspace{-0.4cm}
\caption{Illustration of irregular behaviors captured by \texttt{MRIC}. 
Dynamic visualizations:
\href{https://drive.google.com/file/d/1-G4sv1c84b_77zkiAC0aklTdEAZFbZfV/view?usp=sharing}{U-turn},
\href{https://drive.google.com/file/d/18aw07z5K6Gra6bq8i6MIAPl3blap77Ul/view?usp=sharing}{unprotected left-turn},
\href{https://drive.google.com/file/d/1XzZbG27uNKfLcaxV9GuXEmcReGKgoeHT/view?usp=sharing}{driving onto driveway} and
\href{https://drive.google.com/file/d/1xfdQfHrS3v57b_dCmYPmw4OetgCD2axF/view?usp=sharing}{backing out}.
}
\label{fig:irregular}
\end{figure*}

Fig.~\ref{fig:irregular} shows several real and irregular driving behaviors captured by \texttt{MRIC}, including U-turn, unprotected left-turn, driving onto driveway and backing out of a parking spot. These examples illustrate \texttt{MRIC}'s ability to model complex agent-agent interaction and agent-map interaction in human behaviors.

\subsection{Ablation Study}
We study the effectiveness of the proposed framework component and mixture-of-codebooks module. Considering the high variance of distribution divergence metrics calculated with a small validation set, we present the results on the test set for the ablation study. Note that both the proposed method and the baselines, as described in Sec.~\ref{sec:main_results}, were developed using only the validation set.

\begin{table}[t]
\centering
\caption{Ablation setting on framework components.}
\begin{tabular}{*{5}{c}}
\toprule
Model & \makecell[c]{Open-Loop \\ IL} & \makecell[c]{Closed-Loop \\ IL} & \makecell[c]{Model-Based \\ RL} & \makecell[c]{Dynamic \\ Multipliers} \\
\midrule
$\mathcal{M}_{0}$ &  & \checkmark  &  &   \\
$\mathcal{M}_{1}$ & \checkmark &  &  &      \\
$\mathcal{M}_{2}$ & \checkmark & \checkmark  &  &   \\
$\mathcal{M}_{3}$ & \checkmark & \checkmark  & $\times$(Model-free)  &  \\
$\mathcal{M}_{4}$ & \checkmark & \checkmark  & \checkmark  &  \\
$\mathcal{M}_{*}$ & \checkmark & \checkmark  & \checkmark  &  \checkmark \\
\bottomrule
\end{tabular}
\label{tab:ablation_setting_framework}
\end{table}

\begin{table}[t]
\centering
\caption{Ablation setting on latent space structure. ``State-observation'' denotes using state-observation pair as input of posterior network. Similar for ``State''. ``Fixed" and ``Dynamic" denote the same and various prior distribution parameters under different observations. ``Scenario", ``Step'' and ``Maneuver'' denote different durations of a sampled latent variable.}
\begin{tabular}{*{5}{c}}
\toprule
Model & Posterior & Prior & Horizon & Type \\
\midrule
$\mathcal{M}_{2,0}$ & State & Fixed & Scenario & Gaussian \\
$\mathcal{M}_{2,1}$ & State-observation & Fixed & Scenario & Gaussian \\
$\mathcal{M}_{2,2}$ & State-observation & Fixed & Step  & Gaussian    \\
$\mathcal{M}_{2,3}$ & State-observation & Dynamic & Step & Gaussian    \\
$\mathcal{M}_{2,4}$ & State-observation & Dynamic & Maneuver  & Gaussian  \\
$\mathcal{M}_{2}$ & State-observation & Dynamic & Maneuver  & Codebooks  \\
\bottomrule
\end{tabular}
\label{tab:ablation_setting_latent}
\end{table}

\begin{figure}
\centering
\captionsetup[subfigure]{labelformat=empty} 
\subfloat[]{\includegraphics[]{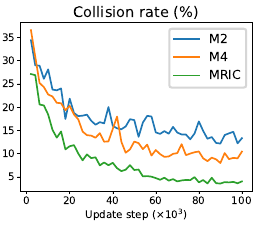}}\label{}\hfil
\subfloat[]{\includegraphics[]{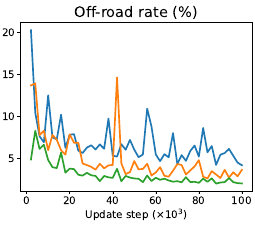}}\label{}\hfil\\
\vspace{-0.7cm}
\subfloat[]{\includegraphics[]{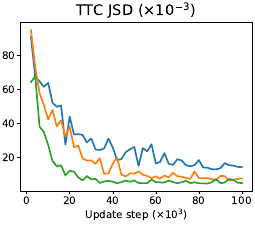}}\label{}\hfil
\subfloat[]{\includegraphics[]{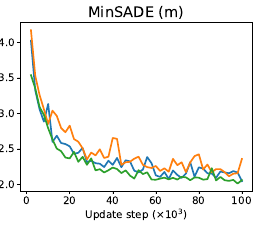}}\label{}\hfil\\
\vspace{-0.5cm}
\caption{Training curves of framework variants. Different from common model-free RL algorithms, our method has much lower variance thanks to the end-to-end differentiability and analytic environment model.}
\label{fig:training_curves}
\end{figure}

\begin{table*}[ht]
\centering
\caption{Ablation study results. ``*'' denotes training does not converge.}
\begin{tabular}{*{15}{c}}
\toprule
 & \multicolumn{3}{c}{Reconstruction (m) $\downarrow$} & \multicolumn{4}{c}{Behavior realism (\%) $\downarrow$} & \multicolumn{7}{c}{Distribution matching JSD ($\times {10}^{-3}$) $\downarrow$} \\
\cmidrule(r){2-4} \cmidrule(lr){5-8} \cmidrule(lr){9-15}
Model & \makecell[c]{Min \\ ADE} & \makecell[c]{Min \\ SADE} & ADE & Coll. & \makecell[c]{Off- \\ road} & \makecell[c]{Rule \\ vio.} & \makecell[c]{Kin. \\ inf.} & Spd. & Acc. & \makecell[c]{Dist.2 \\ obj.} & TTC & \makecell[c]{Dist.2 \\ edge} & Curv. & Prog. \\
\midrule
Log & - & - & - & 3.34 & 1.53 & 3.53 & 1.40 & - & - & - & - & - & - & - \\
\cmidrule(lr){1-15}
$\mathcal{M}_{0}$ & * & * & * & * & * & * & * & * & * & * & * & * & * & * \\
$\mathcal{M}_{1}$ & 7.19 & 8.19 & 9.74 & 41.61 & 35.60 & 5.90 & 0. & 21.32 & 78.86 & 22.14 & 112.24 & 15.53 & 32.03 & 31.89 \\
$\mathcal{M}_{2}$ & 1.17 & 2.02 & 3.73 & 11.36 & 4.38 & 4.23 & 0. & 6.13 & 24.67 & 5.38 & 9.60 & \textbf{1.08} & 5.01 & 1.11 \\
$\mathcal{M}_{3}$ & 3.63 & 5.60 & 9.32 & 49.86 & 49.52 & 4.77 & 0.10 & 74.99 & 67.05 & 34.04 & 100.29 & 34.88 & 85.37 & 12.31 \\
$\mathcal{M}_{4}$ & 1.25 & 2.06 & \textbf{3.65} & 8.02 & 3.37 & 3.77 & 0. & 2.42 & 31.14 & \textbf{2.17} & 2.32 & 1.09 & 2.86 & \textbf{0.69} \\
$\mathcal{M}_{*}$ & \textbf{1.15} & \textbf{1.97} & 3.68 & \textbf{2.56} & \textbf{2.56} & \textbf{3.26} & \textbf{0.} & \textbf{1.92} & \textbf{5.81} & 3.50 & \textbf{1.51} & 4.56 & \textbf{1.30} & 1.21 \\
\cmidrule(lr){1-15}
$\mathcal{M}_{2,0}$ & 2.90 & 3.15 & 3.53 & 8.32 & 5.31 & 3.62 & 0. & 8.25 & 36.52 & \textbf{3.38} & 11.18 & 1.90 & 3.93 & 6.14 \\
$\mathcal{M}_{2,1}$ & 2.09 & 2.49 & \textbf{3.25} & \textbf{7.78} & \textbf{4.02} & 3.66 & 0. & 7.13 & 40.13 & 3.72 & \textbf{6.87} & \textbf{0.81} & 3.56 & 1.54 \\
$\mathcal{M}_{2,2}$ & 1.95 & 3.07 & 5.00 & 18.07 & 4.42 & 6.80 & 0. & 8.40 & 14.46 & 12.32 & 33.68 & 2.04 & 16.27 & 15.34 \\
$\mathcal{M}_{2,3}$ & 1.62 & 2.42 & 3.74 & 11.91 & 5.52 & \textbf{3.60} & 0. & 4.13 & \textbf{11.91} & 5.87 & 12.21 & 1.01 & 9.21 & 2.88 \\
$\mathcal{M}_{2,4}$ & 1.20 & 2.16 & 3.96 & 14.02 & 5.97 & 5.02 & 0. & \textbf{2.71} & 20.90 & 7.15 & 18.74 & 2.10 & \textbf{2.86} & 1.16 \\
$\mathcal{M}_{2}$ & \textbf{1.17} & \textbf{2.02} & 3.73 & 11.36 & 4.38 & 4.23 & \textbf{0.} & 6.13 & 24.67 & 5.38 & 9.60 & 1.08 & 5.01 & \textbf{1.11} \\
\bottomrule
\end{tabular}
\label{tab:ablation_result}
\end{table*}

\subsubsection{Effect of each framework component}\label{sec:ablation_framework}
Table~\ref{tab:ablation_setting_framework} describes the setting of the ablation study for framework components. Table~\ref{tab:ablation_result} shows the results of each framework variant.
As we can see, all the framework components contribute remarkably to the final performance. 
In particular, pure closed-loop IL via differentiable simulation ($\mathcal{M}_{0}$) cannot converge stably due to the gradient exploding issue highlighted in Sec.~\ref{sec:closed-loop}. 
Pure open-loop IL ($\mathcal{M}_{1}$) performs poorly due to covariate shift and causal confusion issues.
Instead, the proposed open-loop model-based IL regularization for closed-loop model-based IL ($\mathcal{M}_{2}$) achieved stable training and much better performance (e.g., minSADE) simultaneously. 
This is attributed to the meaningful learning signals and efficient credit assignment provided by the state-matching via closed-loop differentiable simulation.
However, the pure IL variant ($\mathcal{M}_{2}$) is more likely to generate unrealistic behaviors, such as high collision rates and acceleration JSD, due to weak supervision in low-density regions of the ELBO objective.
$\mathcal{M}_{3}$ combined IL with model-free RL algorithm (i.e., PPO), which led to severe performance degradation. We ascribe this to the high variance of the policy gradient in model-free RL, which negatively interferes with the latent space learning in multi-modal IL.
In contrary, the proposed model-based RL regularization, $\mathcal{M}_{4}$, improved the behavioral realism with significantly less interference with IL's performance. We attribute this difference to the low-variance analytic policy gradient and the elaborately designed differentiable rewards.
However, the effectiveness of applying regularization through simple gradient accumulation remains limited (e.g., collision rate) because the final gradient can be dominated by the main objective.
Finally, the proposed integrated framework $\mathcal{M}_{*}$ not only improved the reconstruction performance, but also significantly enhanced behavioral and distributional realism (e.g., collision rate, off-road rate, and acceleration JSD). 
Combined with the training curves of the variants shown in Fig.~\ref{fig:training_curves}, it is verified that the dynamic multiplier mechanism effectively eliminates the negative interference of open-loop IL and model-based RL regularizations on the closed-loop IL objective, while simultaneously ensuring the effectiveness of these regularizations, leading to a more stable and efficient constrained policy optimization process.

Fig.~\ref{fig:ablation_framework} further compares the generated behaviors in a challenging scenario and the TTC distribution between the pure IL variant $\mathcal{M}_{2}$ and the overall framework $\mathcal{M}_{*}$.
While the pure IL variant produced unrealistic behaviors (collision, driving off-road and running a red light), the integrated framework exhibited safety awareness and successfully accounted for the complex agent-agent and agent-map interaction. This is also reflected in the TTC histogram.

\subsubsection{Effect of latent space structure}\label{sec:ablation_latent}
The structure of the latent space significantly affects both the diversity and stability of the hierarchical framework.
Table~\ref{tab:ablation_setting_latent} presents the ablation study settings for latent space structure, designed in reference to existing works (i.e., $\mathcal{M}_{2,0}$ vs. \texttt{RTC}, $\mathcal{M}_{2,1}$ vs. \texttt{VAE-GAIL}, $\mathcal{M}_{2,2}$ vs. \texttt{ITRA}, $\mathcal{M}_{2,3}$ vs. \texttt{TrafficSim}).
The ablations were conducted on the pure IL variant $\mathcal{M}_{2}$ to emphasize the effect of the latent space.
As shown in Table~\ref{tab:ablation_result}, the proposed mixture-of-codebooks with temporal abstraction mechanism achieves the best overall multi-modal performance.
Concretely, when comparing the result of $\mathcal{M}_{2,0}$ and $\mathcal{M}_{2,1}$, we find that observation information is essential for the posterior network to understand the interaction with neighboring objects and context maps, thereby effectively inferring the unobservable behavioral latents. 
Comparing $\mathcal{M}_{2,1}$ with other ablations reveals that using a single latent variable throughout the entire simulation enhances behavioral realism because it simplifies learning for the low-level policy.
However, it suffers from a lack of diversity, which is critical for multi-modal IL.
On the other hand, using a stepwise changing latent with a fixed prior ($\mathcal{M}_{2,2}$) results in the worst performance due to the simultaneous occurrence of prior holes~\cite{rezende2018taming} and posterior collapses~\cite{lucas2019don} issues.
Employing a dynamic prior distribution ($\mathcal{M}_{2,3}$) alleviates the prior holes problem since the prior network can adjust the latent distribution according to current observation.
Introducing the temporal abstraction mechanism ($\mathcal{M}_{2,4}$) ameliorates the posterior collapse issue by modeling the time scale of human behavior more reasonably.
Finally, using a mixture-of-codebooks further improves behavioral realism by restricting the latent space to avoid unseen samples during execution.
Meanwhile, the reduced reconstruction metrics indicate that the codebook mixtures can effectively compress the dataset and retrieve diverse behavior prototypes of heterogeneous agents.

\section{Conclusion}
In this work, we have proposed \texttt{MRIC}, a model-based reinforcement-imitation learning framework with temporally abstracted mixture-of-codebooks.
In this framework, the meaningful and complementary learning signals for the policy are provided by combining a state-matching-based IL objective with RL regularization.
Efficient credit assignment for high-dimensional inputs over a long horizon is facilitated by closed-loop differentiable simulation and differentiable rewards.
The stable training process is dependent on the open-loop model-based IL regularization.
The main objective's efficient optimization and the regularizations' effectiveness are ensured by a dynamic multiplier mechanism.
The diversity of generated behaviors is attributable to the mixture-of-codebooks module.
In addition, we also reveal the gradient property of closed-loop differentiable simulation, and the nature of the state-matching objective.
Both quantitative and qualitative results show that \texttt{MRIC} is able to simulate diverse and realistic behaviors of heterogeneous agents in various scenarios.
Future work could further extend \texttt{MRIC} to more advanced architectures, such as GPTs, and other autonomous driving tasks, such as planning.

\section*{Appendix}
\renewcommand{\thesubsection}{\Alph{subsection}}

\subsection{Algorithm}
The algorithmic details are provided in Algorithm~\ref{alg:framework}.

\begin{figure*}[!t]
\begin{minipage}{\textwidth}
\begin{algorithm}[H]
\caption{\texttt{MRIC} Algorithm}
\label{alg:framework}
\KwIn{Dataset $\mathcal{D}$, simulation horizon $T$, latent horizon $H$, batch size $B$, number of iteration steps $M$, learning rate $\alpha$ }
\KwOut{High-level policy $\eta_{\theta}$, low-level policy $\pi_{\theta}$, codebooks $E_{\theta}$, posterior network $q_{\phi}$}
Randomly initialize $\theta, \phi$ 

\For{$i=1,\cdots,M$}{
\tcp{Initialize simulation} 
Sample trajectories from dataset: $ \{\tau_j\}_{j=1}^{B} \sim \mathcal{D}$   

Sample environmental latents from empirical distribution: $\xi \sim p(\xi|\tau)$

Sample initial state and observation from empirical distribution: $s_{0}, \bm{o}_{0} \sim p(s_0)p(\bm{o}_{0}|s_0)$

\tcp{Closed-Loop Model-based IL}
\For{$t=0,\cdots,T-1$}{
\eIf{$t \bmod H = 0$}{
Sample behavioral latent from posterior distribution: $\bm{z}_{t} \sim \bm{q}_{\phi}(\bm{z}|\tau)$

Compute prior distribution with generated observation: $\bm{\eta}_{\theta}(\bm{z}|\hat{\bm{o}}_{t})$
}{
Maintain current latent: $\bm{z}_{t} \gets \bm{z}_{t-1}$
}
Take action with generated observation and a posteriori latent: $\hat{\bm{a}}_{t} \sim \bm{\pi}_{\theta}(\bm{a}|\hat{\bm{o}}_{\leq t}, \bm{z}_{t})$

Execute state transition and sample observation from environment: $\hat{s}_{t+1}, \hat{\bm{o}}_{t+1} \sim p(s|\hat{s}_{t}, \hat{\bm{a}}_{t}, u_{t}) p(\bm{o}|\hat{s}_{\leq t+1})$
}
Compute closed-loop IL objective (\ref{elbo_cl}) and acquire gradient w/ BPTT: $\nabla_{\theta, \phi}[\frac{1}{B}\sum_{j=1}^{B}\text{ELBO-CL}(q, \tau_{j}, \xi_{j};\theta, \phi)]$

\tcp{Open-Loop Model-based IL Regularization}
\For{$t=0,\cdots,T-1$}{
\eIf{$t \bmod H = 0$}{
Sample behavioral latent from posterior distribution: $\bm{z}_{t} \sim \bm{q}_{\phi}(\bm{z}|\tau)$

Compute prior distribution with logged observation: $\bm{\eta}_{\theta}(\bm{z}|\bm{o}_{t})$
}{
Maintain current latent: $\bm{z}_{t} \gets \bm{z}_{t-1}$
}
Take action with logged observation and a posteriori latent: $\hat{\bm{a}}_{t} \sim \bm{\pi}_{\theta}(\bm{a}|\bm{o}_{\leq t}, \bm{z}_{t})$

Execute state transition: $\hat{s}_{t+1} \sim p(s|\hat{s}_{t}, \hat{\bm{a}}_{t}, u_{t})$

Sample observation from empirical distribution: $\bm{o}_{t+1} \sim p(\bm{o}|s_{\leq t})$
}
Compute open-loop IL objective (\ref{elbo_ol}) and acquire gradient w/ BPTT: $\nabla_{\theta, \phi}[\frac{1}{B}\sum_{j=1}^{B}\text{ELBO-OL}(q, \tau_{j}, \xi_{j};\theta, \phi)]$

\tcp{Model-Based RL Regularization}
\For{$t=0,\cdots,T-1$}{
\eIf{$t \bmod H = 0$}{
Sample behavioral latent from high-level policy with generated observation: $\hat{\bm{z}}_{t} \sim \bm{\eta}_{\theta}(\bm{z}|\hat{\bm{o}}_{t})$
}{
Maintain current latent: $\hat{\bm{z}}_{t} \gets \hat{\bm{z}}_{t-1}$
}

Take action with generated observation and a priori latent: $\hat{\bm{a}}_{t} \sim \bm{\pi}_{\theta}(\bm{a}|\hat{\bm{o}}_{\leq t}, \hat{\bm{z}}_{t})$ 

Execute state transition and sample observation from environment: $\hat{s}_{t+1}, \hat{\bm{o}}_{t+1} \sim p(s|\hat{s}_{t}, \hat{\bm{a}}_{t}, u_{t}) p(\bm{o}|\hat{s}_{\leq t+1})$
}
Compute model-based RL objective (\ref{objective_rl}) and acquire gradient w/ BPTT: $\nabla_{\theta}[\frac{1}{B}\sum_{j=1}^{B}\mathcal{J}(\hat{\tau}_{j};\theta)]$

\tcp{Solve multiplier \& update parameters}
\For{each parameter group $\mu$}{
Construct gradient matrix: $\hat{G}$

Perform eigenvalue decomposition: $(\Sigma^{2}, V) \gets \text{eig}(\hat{G}^\top\hat{G}) $

Solve multiplier equation (\ref{multiplier_equation}): $\lambda^{*} \gets \sigma V \Sigma^{-1} V^\top \omega$

Update parameter: $\mu \gets \mu - \alpha \hat{G} \lambda^{*}$
}
}
\end{algorithm}
\end{minipage}
\end{figure*}

\subsection{Implementation details}

\subsubsection{Observation}
There are four modalities of data in the observation for each agent, where each modality is represented as a set of elements (e.g., point or state) with specific features.
Table~\ref{tab:observation} provides the details of observation.

\begin{table}[ht]
\centering
\caption{Observation descriptions.}
\begin{tabular}{*{3}{c}}
\toprule
Modality & \#Elements & Features \\
\midrule
Ego trajectory & 6 & \multirow{2}{*}{Position, yaw, speed, velocity, extent}  \\
Object state & 16 &  \\
Static map & 2000 & Position, direction \\
Traffic light & 16 & Position, state \\
\bottomrule
\end{tabular}
\label{tab:observation}
\end{table}

\subsubsection{Reward}
Table~\ref{tab:reward} lists the thresholds used in the differentiable rewards. 

\begin{table}[ht]
\centering
\caption{Thresholds in rewards.}
\begin{tabular}{*{3}{c}}
\toprule
Reward & Lower & Upper \\
\midrule
Collision avoidance & $-\infty$ & $1$  \\
On-road & $-1$ & $+\infty$ \\
Traffic rule compliance & $0$ & $2$ \\
\bottomrule
\end{tabular}
\label{tab:reward}
\end{table}

\subsubsection{Network architecture}
Table~\ref{tab:net} lists the key hyperparameters that defined the neural networks. 
Table~\ref{tab:parameter} shows the number of parameters of each network. 

\begin{table}[ht]
\centering
\caption{Network hyperparameters.}
\begin{tabular}{*{3}{c}}
\toprule
Module & Hyperparameter & value \\
\midrule
\multirow{4}{*}{TokenNet} & encode\_dim & 32  \\
& \#queries & 1 \\
& query\_dim & 128 \\
& \#cross\_attn\_heads & 4 \\
\cmidrule(lr){1-3}
\multirow{2}{*}{Codebook} & \#embedding\_vectors & 128  \\
& embed\_dim & 128 \\
\cmidrule(lr){1-3}
\multirow{2}{*}{Policy} & gru\_hidden\_size & 128  \\
& mlp\_hidden\_size & 128 \\
\cmidrule(lr){1-3}
Posterior & \#self\_attn\_heads & 4 \\
\bottomrule
\end{tabular}
\label{tab:net}
\end{table}

\begin{table}[ht]
\centering
\caption{Network size.}
\begin{tabular}{*{2}{c}}
\toprule
Network & \#Params \\
\midrule
High-level policy & $811,744$  \\
Low-level policy w/ codebooks & $950,695$ \\
Posterior & $1,010,016$ \\
\bottomrule
\end{tabular}
\label{tab:parameter}
\end{table}

\subsubsection{Simulation}
Table~\ref{tab:simulation} lists the hyperparameters that defined the simulation process in training and execution. 

\begin{table}[ht]
\centering
\caption{Simulation hyperparameters.}
\begin{tabular}{*{3}{c}}
\toprule
Hyperparameter & value & Remark \\
\midrule
\#Simulation steps $T$ & 40 &  \\
\#Initialization steps & 6 &  \\
Frequency & $5Hz$ & \\
Map resolution & $1m$ & \\
Max \#agents & 256 & \\
Max \#lane points & $10K$ & \\
Rear axle distance coefficient $c_{r}$ & 0.3 & \multirow{4}{*}{Bicycle motion model} \\
Front axle distance coefficient $c_{f}$ & 0.3 &  \\
Max acceleration & $6 m/s^{2}$ & \\
Max steering angle & $45^{\circ}$ & \\
Behavioral latent duration $H$ & 5 steps & Temporal abstraction \\
Discount factor $\gamma$ & 1.0 & POSG \\
\bottomrule
\end{tabular}
\label{tab:simulation}
\end{table}

\subsubsection{Training}
We trained the neural networks with two NVIDIA RTX 3090 GPUs, which takes about 5 days to complete.
The training checkpoint with lowest minSADE was selected for evaluation, which takes about 8 hours to finish.
Table~\ref{tab:train} lists the hyperparameters used for training.
In practise, we implemented the log-likelihood as Huber loss for position and mean square error for heading angle.

\begin{table}[ht]
\centering
\caption{Training hyperparameters.}
\begin{tabular}{*{2}{c}}
\toprule
Hyperparameter & value \\
\midrule
Batch size $B$ & 16  \\
\#Update steps $M$ & $100K$ \\
Validation intervals & $2K$ \\
Learning rate $\alpha$ & 1e-4 \\
Optimizer & AdamW \\
Gradient clipping (norm) & 1.0 \\
Regularization weight $\omega$ & $(0.6, 0.3, 0.1)^{\text{T}}$ \\
\bottomrule
\end{tabular}
\label{tab:train}
\end{table}

\subsubsection{Histograms}
Table~\ref{tab:histogram} lists the hyperparameters that defined the histograms of state features and trajectory features.

\begin{table}[ht]
\centering
\caption{Histograms hyperparameters.}
\begin{tabular}{*{4}{c}}
\toprule
Histogram & Min & Max & \#Bins  \\
\midrule
Linear speed $(\mathrm{m/s})$ & $0$ & $35$ & \multirow{9}{*}{$200$}  \\
Angular speed $(\mathrm{rad/s})$ & $-1$ & $1$ & \\
Linear acceleration $(\mathrm{m/s^{2}})$ & $-10$ & $10$ &  \\
Angular acceleration $(\mathrm{rad/s^{2}})$ & $-2$ & $2$ & \\
Distance to nearest object $(\mathrm{m})$ & $-5$ & $40$ & \\
Time-to-collision $(\mathrm{s})$ & $0$ & $5$ & \\
Distance to road edge $(\mathrm{m})$ & $-20$ & $40$ & \\
Curvature $(\mathrm{m^{-1}})$ & $-0.2$ & $0.2$ & \\
Progress $(\mathrm{m})$ & $0$ & $280$ & \\
\bottomrule
\end{tabular}
\label{tab:histogram}
\end{table}

\subsubsection{Search algorithm}
Table~\ref{tab:dfs} lists the hyperparameters used in the depth-first-search algorithm.

\begin{table}[ht]
\centering
\caption{DFS hyperparameters.}
\begin{tabular}{*{2}{c}}
\toprule
Hyperparameter & value \\
\midrule
\#Route candidates & 64  \\
Search depth & 180 \\
\bottomrule
\end{tabular}
\label{tab:dfs}
\end{table}

\bibliographystyle{IEEEtran}
\bibliography{References}

\end{document}